%% file: ProbRobustSSI.tex
\documentclass[preprint,12pt,sort&compress]{elsarticle}
\usepackage[english]{babel} 
\usepackage{graphicx}
\usepackage{amsthm,amsmath,amssymb}
\usepackage{float}
\usepackage{amsfonts}
\usepackage{a4wide}
\usepackage{microtype}
\usepackage{booktabs}
\usepackage[modulo]{lineno}
\usepackage{circuitikz}
\usepackage[edges]{forest}
\usepackage{array}
\usetikzlibrary{shapes.gates.logic.US,trees,positioning,arrows,shapes.geometric}
\usetikzlibrary{calc,patterns,trees,decorations.pathmorphing,decorations.markings}

\usepackage{rotating}
\usepackage{tikz}
\usepackage{pgfplots}
\usetikzlibrary{bayesnet}
\usetikzlibrary{arrows}

\usepackage{orcidlink}
\input{drg_continuity.tex}

\usepackage{hyperref} 

\journal{JSV}

\begin{document}

\begin{frontmatter}

	\title{A Robust Probabilistic Approach to Stochastic Subspace Identification} 

	\author[1]{B. J. O'Connell \orcidlink{0000-0001-6042-927X} \corref{cor1}}
	\ead{bjoconnell1@sheffield.ac.uk}
	\address[1]{Dynamics Research Group, Department of Mechanical Engineering, The University of Sheffield, Mappin Street, Sheffield, S1 3JD, UK}
	\cortext[cor1]{Corresponding Author}
	\author[1]{T. J. Rogers \orcidlink{0000-0002-3433-3247}}
	
	\begin{abstract}

		\noindent Modal parameter estimation of operational structures is often a challenging task when confronted with unwanted distortions (outliers) in field measurements. Atypical observations present a problem to operational modal analysis (OMA) algorithms, such as stochastic subspace identification (SSI), severely biasing parameter estimates and resulting in misidentification of the system. Despite this predicament, no simple mechanism currently exists capable of dealing with such anomalies in SSI. Addressing this problem, this paper first introduces a novel probabilistic formulation of stochastic subspace identification (Prob-SSI), realised using probabilistic projections. Mathematically, the equivalence between this model and the classic algorithm is demonstrated. This fresh perspective, viewing SSI as a problem in probabilistic inference, lays the necessary mathematical foundation to enable a plethora of new, more sophisticated OMA approaches. To this end, a statistically robust SSI algorithm (robust Prob-SSI) is developed, capable of providing a principled and automatic way of handling outlying or anomalous data in the measured timeseries, such as may occur in field recordings, e.g. intermittent sensor dropout. Robust Prob-SSI is shown to outperform conventional SSI when confronted with `corrupted' data, exhibiting improved identification performance and higher levels of confidence in the found poles when viewing consistency (stabilisation) diagrams. Similar benefits are also demonstrated on the Z24 Bridge benchmark dataset, highlighting enhanced performance on measured systems.

	\end{abstract}
	
	
\end{frontmatter}



\section{Introduction}\label{sec:Introduction}

The characterisation of structural dynamic systems remains to be a key feature of modern engineering practice. Modal analysis is an established field of system identification that is frequently employed across industry and research to practically analyse (linear) dynamic systems with a view to recover estimates for the underlying invariant (modal) properties. These are familiar to the dynamicist as the natural frequencies, damping ratios and mode shapes. The availability of this modal information helps facilitate more informed decision-making throughout the entire lifecycle of engineering structures.

\begin{figure}[H]
	\centering
	\includegraphics[height=0.8\textheight]{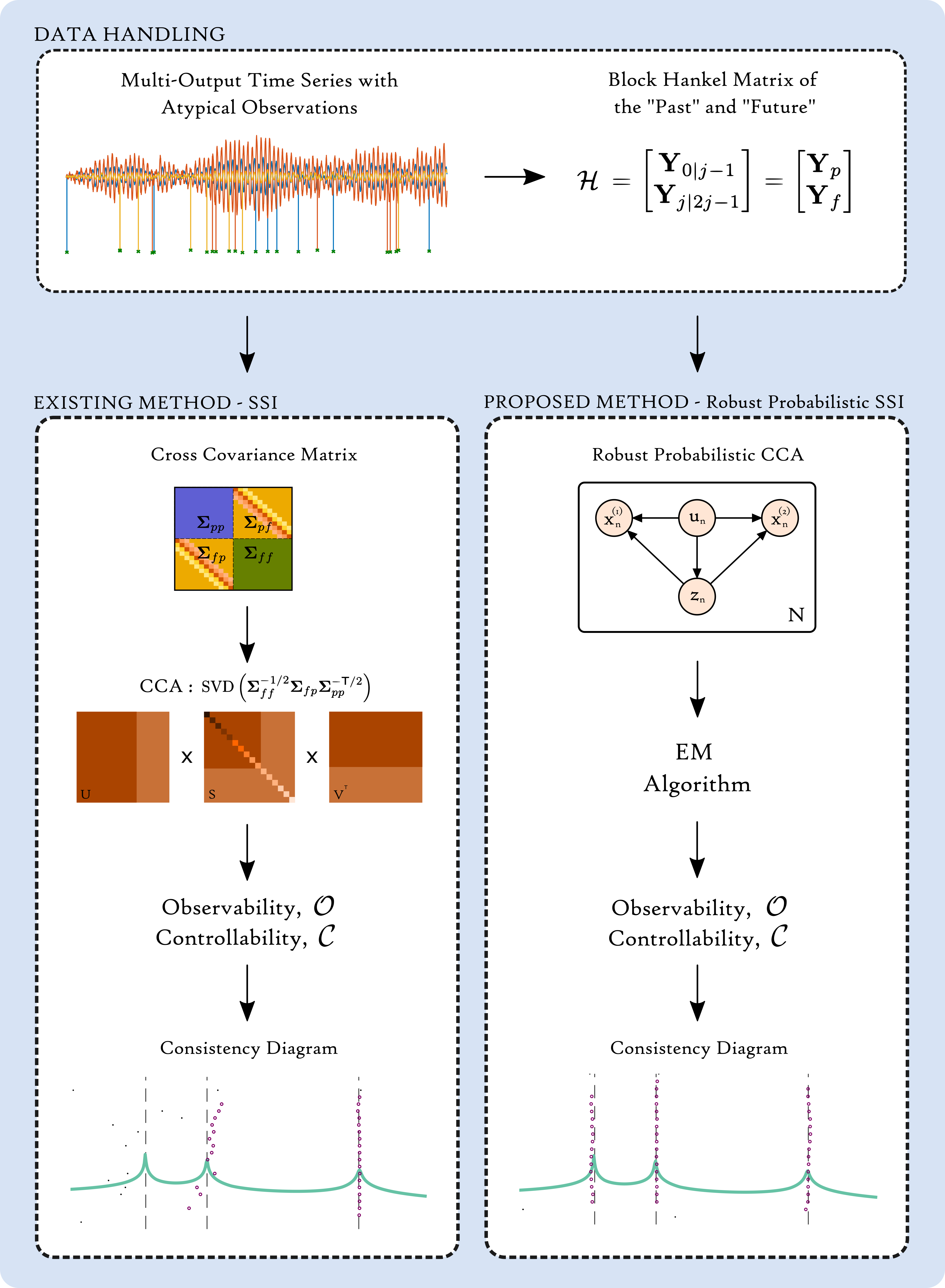}
	\caption{Schematic of the the existing methodology, SSI, and the proposed methodology, Robust Prob-SSI, including an example of the observed relative improvement to identification performance from robust Prob-SSI.}
	\label{fig:schematic}
\end{figure}

Subspace identification algorithms have gained increasing attention in the modal analysis community \cite{Peeters2001}, predominantly due to their ability to deal with a considerable number of inputs and outputs. Stochastic subspace identification (SSI) is a prominent method capable of estimating a linear time-invariant state-space model from correlated sequences of observed data, using traditional linear algebra techniques. Once recovered, an eigen-decomposition of the state-space model yields the desired modal properties. Frequently appearing in both academic and industrial literature, SSI (and its many algorithmic variants \cite{Peeters1999b,James1993}) continues to be employed as reliable means of operational modal analysis (OMA) \cite{Peeters2001,Reynders2012a,Loh2011,Nord2019a,Jin2021}. In this particularly paper, we focus attention to covariance-driven stochastic subspace identification (Cov-SSI). Schematically, the Cov-SSI procedure is shown on the left hand side of Figure \ref{fig:schematic} and described in full in Section \ref{sec:CovSSI}.

Despite the success of OMA algorithms, a typical weakness lies in their handling of atypical observations and non-Gaussian noise. In practice, in situ monitoring can often produce imperfect data containing unwanted features that may present a problem to OMA algorithms, severely biasing parameter estimates and inevitably leading to the misidentification of a system. These features arise due to the low level of excitation, the inherent stochastic nature of the forcing or from unpredicted events that are often independent of the system being measured. In the case of OMA, these `events' refer to practically encountered scenarios in testing such as sensor drop-out or partial detachment.

This misidentification problem is also true of Cov-SSI. Unlike classical approaches to data analysis, where data may be pre-processed to remove outliers and `cleaned' prior to use, such an approach cannot easily be applied to Cov-SSI due to its dependency on sequential data, as required for the Hankel matrix formation. This poses a significant dilemma. During application of SSI, any outliers remain present during the analysis, distorting the measurement of the response and ultimately affecting the identification procedure. Despite this discernible predicament, no simple mechanism currently exists capable of dealing with such anomalies in SSI. 

Addressing this problem, this paper introduces a novel probabilistic formulation of SSI, referred to as Prob-SSI, realised using the theory of probabilistic projections \cite{Bach2005} and latent variable models. This new formulation is achieved through substitution of canonical correlation analysis (CCA) (i.e. the singular value decomposition (SVD)) with probabilistic CCA \cite{Bach2005} within the SSI procedure. This fresh perspective, viewing SSI as a problem in probabilistic inference, now lays the appropriate mathematical foundation to enable a plethora of new SSI-based OMA algorithms, constructed using more sophisticated probabilistic and hierarchical techniques. 

To that end, the authors also present the first of many new algorithms constructed on this unique model - a statistically robust SSI algorithm (robust Prob-SSI) capable of providing a principled and automatic way of handling a-typical observations in multi-output timeseries responses. As will be demonstrated, when confronted with `corrupted` data, this new approach to modal identification appears to outperform traditional SSI, exhibiting improved pole consistency and increased confidence; attributes reflected in the subsequent consistency diagrams. This new proposed methodology is shown on the right hand side of Figure \ref{fig:schematic} and discussed in Section \ref{sec:RobustSSI}. Further investigation and application to the well-documented Z24 Bridge benchmark is also conducted, with results highlighting similar benefits to identification performance.

\subsection{Related Work}


The application of probabilistic methods in modern engineering practice, coupled with the demonstrative value associated with quantifying and harnessing uncertainty through probabilistic analysis, has led to a general shift across many engineering disciplines to adopt probabilistic tools and operate within more probabilistic frameworks. In the realm of structural dynamics, this is most evident in the areas of SHM \cite{Farrar2012}, digital twins \cite{Worden2020} and more increasingly in the field of system identification. The ability to reformulate engineering operations probabilistically not only simplifies their inclusion into probabilistic frameworks but can often provide an opportunity to improve existing methods, by means previously unattainable using a deterministic approach.

Probabilistic (and Bayesian) approaches to modal analysis are now becoming increasingly popular within the dynamics community \cite{Rogers2019}. The inclusion of uncertainty, in its various forms\footnote{It is important to note, that there are many types of uncertainty, whether that be a posterior uncertainty in a Bayesian sense, estimates for bounds, confidence intervals, fiducial intervals, etc. The discussion of their relative merits lies beyond this work.}, has proven increasing useful; aiding the automatic selection of consistent poles in stabilisation diagrams \cite{Qin2016,Priou2022}, quantifying the uncertainty over the modal properties \cite{Reynders2008,Reynders2016,Reynders2021,Pintelon2007} and by providing an alternative foundation upon which new methods can be developed \cite{Bi2018}. 

Despite many alternative approaches to OMA and an increase in probabilistic identification techniques, few directly attempt to formulate statistically robust algorithms capable of handling atypical observations. Most OMA algorithms possess some level of inherent robustness given their dependency on certain mathematical operations, such as the singular value decomposition, but in certain cases, as will be shown, these algorithms are incapable of dealing with a considerable number of outliers. 

Some `robust' approaches to OMA do exist in literature, however, the term `robust' is often used to loosely refer to methods capable of reliable performance, rather than statistical robustness. Most of these methods achieve consistency through automated process; using optimised metrics or machine learning \cite{Cheema2021}, often with the inclusion of uncertainty in techniques such as clustering \cite{Priou2022}. In contrast, this work achieves statistical robustness through direct augmentation of the underlying algorithm. Furthermore, upon closer inspection of the literature, only a small selection of papers appear to directly address the problem of outliers, including recently developed robust algorithms by Liu et.al. on correlation signal subset-based SSI (CoS-SSI) \cite{Liu2019a} and Goursat et.al. \cite{Goursat2011} on Crystal Clear SSI (CC-SSI); both concerned with addressing non-stationary and noisy signals. 

\subsection{Structure}

To assist navigation through the remainder of the article, the structure is given as follows: 
\begin{itemize}
	\item (\textbf{Section 2}) provides the necessary theoretical background for canonical correlation analysis (\textbf{2.1}) and covariance-driven stochastic subspace identification (\textbf{2.2}).
	\item (\textbf{Section 3}) introduces the theory of probabilistic canonical correlation analysis (\textbf{3.1}). This is followed by a definition of the novel probabilistic stochastic subspace identification algorithm (\textbf{3.2}).
	\item (\textbf{Section 4}) introduces the existing theory of robust probabilistic canonical correlation analysis (\textbf{4.1}), including the necessary expectation-maximisation update equations (\textbf{4.1.1}), followed by a definition of the new robust probabilistic stochastic subspace identification algorithm (\textbf{4.2}).
	\item (\textbf{Section 5}) presents results and discussion from three individual case studies: a `clean' benchmark case (\textbf{5.1}), a `corrupted' case containing atypical observations (\textbf{5.2}) and a real world case, using the Z24 bridge benchmark (\textbf{5.3}).
	\item (\textbf{Section 6}) provides concluding remarks and an outward look at possible avenues for future work. 
\end{itemize}



\section{Theoretical background}

\subsection{Canonical Correlation Analysis}\label{sec:CCA}

Canonical correlation analysis (CCA) is a well established statistical tool \cite{Hotelling1936} that is used to analyse the mutual dependency between two sets of multidimensional variables. This dependency can be quantified by finding an appropriate set of orthogonal basis vectors, ${\vec{a}}$ and ${\vec{b}}$, such that the correlation between the linear projections\footnote{By assuming linear projections, the results of CCA are easily interpretable whilst keeping the problems of overfitting mostly manageable.} of the variables is mutually maximised, subject to the constraint that the set of transformed variables are uncorrelated\footnote{One intuitive explanation of CCA is that it is performing PCA on two datasets \(\vec{x}\) and \(\vec{y}\) whilst simultaneously attempting to maximise the correlation of the two sets of principal components.}. Given two co-occurring multidimensional random variables, $\boldsymbol{x} \elmtreal{d_1}$ and $\boldsymbol{y} \elmtreal{d_2}$, this process can be represented mathematically as
\begin{equation}
	({\vec{a}^{\prime}}, {\vec{b}^{\prime}}) \ = \ \argmax{{\vec{a}},{\vec{b}}} \ \ \mathrm{corr}(\vec{a}^{\trans}\boldsymbol{x},\vec{b}^{\trans}\boldsymbol{y}) \ = \ \argmax{{\vec{a}},{\vec{b}}} \ \dfrac{\vec{a}^{\trans}\mat{\Sigma}_{xy}\vec{b}}{\sqrt{\vec{a}^{\trans}\mat{\Sigma}_{xx}\vec{a}\vec{b}^{\trans}\mat{\Sigma}_{yy}\vec{b}^{\trans}}}
\end{equation}
where $\vec{\Sigma}_{xy}$ is defined as the cross covariance matrix between $\boldsymbol{x}$ and $\boldsymbol{y}$, with $\vec{\Sigma}_{xy} = \vec{\Sigma}_{yx}^{\trans}$, and $\vec{\Sigma}_{xx}$ and $\vec{\Sigma}_{yy}$ are the auto covariance matrices. The canonical correlations can be computed easily by solving the generalised eigenvalue problem
\begin{equation}
    \begin{pmatrix}
        0 & \vec{\Sigma}_{xy} \\
        \vec{\Sigma}_{yx} & 0
    \end{pmatrix} 
    \begin{pmatrix}
        \vec{a}\\
        \vec{b} 
    \end{pmatrix} = \lambda
    \begin{pmatrix}
        \vec{\Sigma}_{xx} & 0 \\
        0 & \vec{\Sigma}_{yy}
    \end{pmatrix}
    \begin{pmatrix}
        \vec{a} \\
        \vec{b} 
    \end{pmatrix}
\end{equation}
where $\vec{a}$ and $\vec{b}$ are the eigenvectors and $\lambda$ is the eigenvalue or canonical correlation. In practice the set of eigenvalues and eigenvectors are computed using the singular value decomposition (SVD) of $\vec{\Sigma}_{xx}^{-\frac{1}{2}}\vec{\Sigma}_{xy}\vec{\Sigma}_{yy}^{-\frac{\trans}{2}} = \mat{V}_1 \mat{\Lambda} \mat{V}_2^{\trans}$.

\subsection{Covariance-Driven Stochastic Subspace Identification (Cov-SSI)}\label{sec:CovSSI}

This section provides a concise overview of the theory underpinning Cov-SSI for an output-only case. For the benefit of the reader, the subtle presence of CCA in the method is highlighted. This derivation follows descriptions given by Katayama \cite{Katayama2005}, Van Overschee and De Moor \cite{VanOverscheeDeMoor1996} and D\"{o}hler and Mevel \cite{Dohler2013}. The reader is directed towards the aforementioned texts for a fuller derivation and for specifics. Only important details are repeated here. For ease, Cov-SSI will be synonymous with the abbreviation `SSI' if used, throughout the remainder of the paper.  
	
Consider an $r^{\mathrm{th}}$ order discrete state space model of a linear dynamic system, equivalent to a mechanical system with $n_{\mathrm{dof}}$ degrees of freedom, such that $r = 2 n_{\mathrm{dof}}$, in the form

\begin{equation}
	\begin{aligned}
	\vec{x}_{k+1} &= \vec{A}_d \vec{x}_k + w_k \\
	\vec{y}_k &= \vec{C} \vec{x}_k + v_k
	\end{aligned}
	\quad , \quad
	\mathbb{E} \left\{ \begin{bmatrix} w_q \\ v_q \end{bmatrix} \begin{bmatrix} w^{\trans}_s & v^{\trans}_s \end{bmatrix} \right\} = \begin{bmatrix} \vec{Q} & \vec{S} \\ \vec{S}^{\trans} & \vec{R} \end{bmatrix} \delta_{qs}
	\label{eq: State Space Model}
\end{equation}

where $\vec{y}_k \in \mathbb{R}^{l}$ is the output vector at discrete time step $k$, $\vec{x}_k \in \mathbb{R}^{p}$ is the internal state vector, $\vec{A}_d \in \mathbb{R}^{p \times p}$ is the discrete state matrix such that $\vec{A}_d = \mathrm{exp}(\vec{A}_c \Delta t)$ where $\vec{A}_c$ is the continuous state matrix, $\Delta t$ is the sampling time and $\vec{C}$ is the output matrix. $w_k \in \mathbb{R}^{p}$ and $v_k \in \mathbb{R}^{l}$ are samples of the process noise and measurement noise respectively, $\mathbb{E}[\cdot]$ denotes the expectation and $\delta_{qs}$ is the Kronecker delta for any two samples in time $q$ and $s$. The process and measurement noise are both assumed to be stationary, white noise Gaussian with zero mean and covariance given by the second part of Equation \ref{eq: State Space Model}.

\begin{figure}[h]
	\centering
	\includegraphics[width=\textwidth]{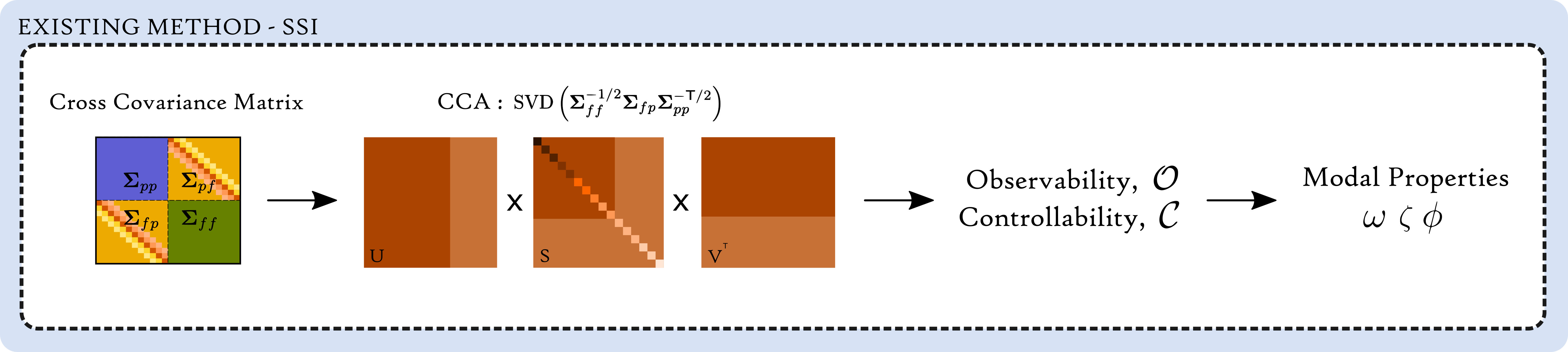}
	\caption{Schematic of the SSI procedure, highlighting the presence of CCA.}
	\label{fig:SSI Flowchart}
\end{figure}

	
Given output measurements from this stationary process, based on all $l$ measurement channels, the data can be arranged into a block Hankel matrix, in the usual way, to form     

\begin{equation}
   \mathcal{H} = \vec{Y}_{0|2j-1}  \in \mathbb{R}^{2lj \times N} \ \ =  \ \ \begin{bmatrix} \vec{Y}_{0|j-1} \\ \vec{Y}_{j|2j-1}\end{bmatrix} \ \ = \ \ \begin{bmatrix} \vec{Y}_p \\ \vec{Y}_f \end{bmatrix} 
\end{equation}

\noindent with $2j$ block rows and $N$ columns, with every block consisting of $l$ rows and where $j>0$ and $N$ is sufficiently large (i.e much larger than $2lj$) and where $j > r$ (model order) and the number of columns of block matrices is $N$. The resultant cross-covariance matrix of the future $\vec{Y}_f$ with the past $\vec{Y}_p$ is therefore given by

\begin{equation}
	\tilde{\mat{\Sigma}} \ \ = \ \ \dfrac{1}{N}\begin{bmatrix} \vec{Y}_p \\ \vec{Y}_f \end{bmatrix} \begin{bmatrix} \vec{Y}_p^{\trans} & \vec{Y}_f^{\trans} \end{bmatrix} \ \ = \ \ \begin{bmatrix} \vec{\Sigma}_{pp} & \vec{\Sigma}_{pf} \\ \vec{\Sigma}_{fp} & \vec{\Sigma}_{ff} \end{bmatrix} 
\end{equation}
    
\noindent where $\vec{\Sigma}_{pf}$ and $\vec{\Sigma}_{fp}$ are block cross-covariance matrices, and $\vec{\Sigma}_{ff}$, $\vec{\Sigma}_{pp}$ are block auto-covariance matrices respectively. The canonical correlations $\vec{\Lambda} = \mathrm{diag}(\lambda_1, \cdots, \lambda_r)$ between the future and past are the singular values \cite{Katayama2005}, obtained through the singular value decomposition (SVD) of the following matrix

	\begin{equation}
		\vec{\Sigma}_{ff}^{-1/2}\vec{\Sigma}_{fp}\vec{\Sigma}_{pp}^{-\trans/2} \ \ = \ \ \vec{V}_1 \vec{\Lambda} \vec{V}_2^{\trans} \ \ \backsimeq \ \ \breve{\vec{V}}_1 \breve{\vec{\Lambda}} \breve{\vec{V}}_2^{\trans}
		\label{eq:Normalised Covariance 1}
	\end{equation}

\noindent where $\vec{\Sigma}_{ff}^{1/2}\vec{\Sigma}_{ff}^{\trans/2} = \vec{\Sigma}_{ff}$, such that, 

	\begin{equation}
		\vec{\Sigma}_{fp} \ \ \backsimeq \ \  \vec{\Sigma}_{ff}^{1/2} \breve{\vec{V}}_1 \breve{\vec{\Lambda}} \breve{\vec{V}}_2^{\trans} \vec{\Sigma}_{pp}^{\trans/2}
		\label{eq:Normalised Covariance 2}
	\end{equation}

\noindent where $\vec{V_1}$ and $\vec{V}_2$ are the left and right singular vectors, respectively and $\breve{\vec{\Lambda}}$ neglects sufficiently small singular values (canonical correlations) in $\vec{\Lambda}$ such that the resultant state vector has the dimension $d = \mathrm{dim}(\breve{\vec{\Lambda}})$. The cross-covariance matrix, $\vec{\Sigma}_{pf}$, can be decomposed into the corresponding extended observability $(\mathcal{O})$ and controllability $(\mathcal{C})$ matrices using $\vec{\Sigma}_{fp} = \mathcal{O}\mathcal{C}$ where
	\begin{equation}
		\mathcal{O}  \ \ = \ \ \vec{\Sigma}_{ff}^{1/2}\breve{\vec{V}}_1\breve{\vec{\Lambda}}^{1/2} \ \ \ , \ \ \ \ \mathcal{C} \ \ = \ \ \breve{\vec{\Lambda}}^{1/2}\breve{\vec{V}}_2^{\trans}\vec{\Sigma}_{pp}^{\trans/2}
		\label{eq:ObservReach}
	\end{equation}

respectively and where rank($\mathcal{O}$) = rank($\mathcal{C}$) = $d$. The extended controllability and observability matrices can be used to recover the state $\vec{A}$ and output $\vec{C}$ matrices,\footnote{The method described here for recovering the system matrices is one of many. See Stochastic Balanced Realisation Algorithm B, Chapter 8 in Katayama \cite{Katayama2005}} and therefore the modal properties, in the usual manner for operational modal analysis (see \ref{A: Recovery of ModalProps}). This overall process is summarised by the schema shown in Figure \ref{fig:SSI Flowchart}.

\section{Probabilistic Stochastic Subspace Identification}\label{sec:ProbSSI}

\subsection{Probabilistic CCA (PCCA)} \label{subsec:PCCA}

The probabilistic interpretation of CCA, developed by Bach and Jordan \cite{Bach2005}, uses a latent model approach, similar to that applied by Tipping and Bishop \cite{TippingBishop1999} in their earlier development of probabilistic principal component analysis. Probabilistic CCA is summarised here and, for the remainder of the paper, will be abbreviated to PCCA. The term CCA will solely refer to the traditional CCA method.

\begin{figure}[h]
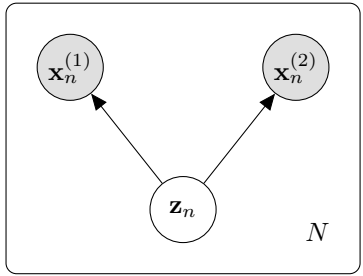

	\centering
	  \tikz{
		\tikzstyle{latent} = [circle,fill=white,draw=black,inner sep=1pt,
		minimum size=25pt, font=\fontsize{10}{10}\selectfont, node distance=1]
		 \node[latent] (z) {$\vec{z}_n$};%
		  \node[obs,above=of z, xshift=-1.5cm] (x1) {$\vec{x}_n^{(1)}$}; %
		  \node[obs,above=of z, xshift=1.5cm] (x2) {$\vec{x}_n^{(2)}$}; %
		\tikzstyle{plate caption} = [caption, node distance=0, inner sep=0pt,
			 below left=-8pt and 0pt of #1.south east]
		  \plate [inner sep=0.4cm] {plate1} {(z)(x1)(x2)} {$N$}; %
		  \edge {z} {x1,x2}  
		  } 
	 \caption{Graphical model for the probabilistic, latent variable interpretation of CCA (PCCA)}
	 \label{fig:Graph PCCA}
\end{figure}

	At the core of PCCA lies a lower dimensional, unobserved latent space described by variable $\zlatentn{n} \elmtreal{d}$. This latent variable is transformed through two independent linear mappings\footnote{Note that the transformations defined by Bach and Jordan transform variables from the latent space to the data space.}, $\wght{(1)}$ and $\wght{(2)}$, to produce a pair of observed variables $\xdatan{n}{(1)} \elmtreal{D_1}$ and $\xdatan{n}{(2)} \elmtreal{D_2}$.  Crucially, as the observed variables are assumed to be generated from a shared latent space, the two datasets can be considered correlated. The full set of observed samples are given by matrices $\mat{X}^{(m)} = [\xdatan{1}{(m)},\hdots,\xdatan{N}{(m)}] \elmtreal{D_m \times N}$ where $m = 1,2$ and $N$ is the total number of observations. To simplify notation, $\xdatan{n}{} = [\xdatan{n}{(1)};\xdatan{n}{(2)}] \elmtreal{D}$ and $\mat{X} = [\mat{X}^{(1)};\mat{X}^{(2)}] \elmtreal{D \times N}$ is the feature-wise row concatenation of the two random variables, where $D = D_1 + D_2$. PCCA is parameterised by the model parameters $\mat{\theta} = \{{\wght{},\covar{},\mean{},\mat{Z}}\}$ where the latent variables are denoted by $\mat{Z} = [\zlatentn{1}, \hdots, \zlatentn{N}]$, one for each pair of observed samples, and the remaining variables are the independent model parameters. Given this list of assumptions, Bach and Jordan defined the following probabilistic model, shown graphically in Figure \ref{fig:Graph PCCA} and given mathematically through Equations \ref{eq: distr latent z} - \ref{eq: distr x}.
	\begin{align}
		\vec{z}_n \ \ &\sim \ \ \mathcal{N}(0,\ident) \label{eq: distr latent z}\\
		\vec{x}^{(m)}_n | \vec{z}_n \ \ &\sim \ \ \mathcal{N}(\vec{\mathrm{W}}^{(m)}\vec{z}_n + \vec{\mu}^{(m)},\vec{\Sigma}^{(m)}) \label{eq: distr x(m)}\\
		\vec{x}_n | \vec{z}_n \ \ &\sim \ \ \mathcal{N}(\vec{\mathrm{W}}\vec{z}_n + \vec{\mu},\vec{\Sigma}) \label{eq: distr x}
	\end{align}
	where $\gaussdistr{\mean{}}{\covar{}}$ corresponds to a Gaussian distribution characterised by mean and covariance, $\wght{} = [\wght{(1)};\wght{(2)}]$, $\mean{} = [\mean{(1)};\mean{(2)}]$ and $\covar{}$ is a block-diagonal covariance matrix with $\covar{(1)}$ and $\covar{(2)}$ along the diagonal.
	An isotropic noise model is assumed in the latent space, which enforces independence between the variables whilst imposing a maximum correlation condition.
	
	\subsubsection{Maximum Likelihood Estimates}\label{sec: MLE PCCA}
	Using this model, Bach and Jordan demonstrated the equivalence of probabilistic CCA to traditional CCA by proving that the maximum likelihood estimates (MLE) for the parameters in Equation \ref{eq: distr x(m)} lead to the results of classical CCA and therefore, contain all the necessary information traditionally obtained through the SVD. The MLE of the weights are given by
	\begin{align}
		\wghtmle{(1)} \ \ &= \ \ {\covar{1/2}_{11}}\mat{V}_{1}\mat{P}^{1/2}\mat{R}  \\
		\wghtmle{(2)} \ \ &= \ \ {\covar{1/2}_{22}}\mat{V}_{2}\mat{P}^{1/2}\mat{R}
	\end{align}	
	where $\mat{V}_1$ and $\mat{V}_2$ are the left and right singular vectors of $\vec{\Sigma}_{11}^{-1/2}\vec{\Sigma}_{12}\vec{\Sigma}_{22}^{-\trans/2}$, respectively, $\mat{P}$ is a diagonal matrix of the largest $d$ canonical correlations and $\mat{R}$ is an arbitrary rotation matrix of size $d$. The MLE of the means $\mu_1$ and $\mu_2$ are not stated here as the assumption is zero mean data but their addition, if required, is trivial.

	\subsubsection{Expectation-Maximisation Algorithm}

	A convenient property of latent variable models is that model parameter estimates can also be recovered iteratively using the expectation-maximisation (EM) algorithm \cite{Dempster1977}. Despite fully tractable MLE estimates, Bach and Jordan also provide EM update equations for PCCA (see Section 4.1 in \cite{Bach2005}). 
	

\subsection{Probabilistic SSI}

	\begin{figure}[h]
		\centering
		\includegraphics[width=0.98\textwidth]{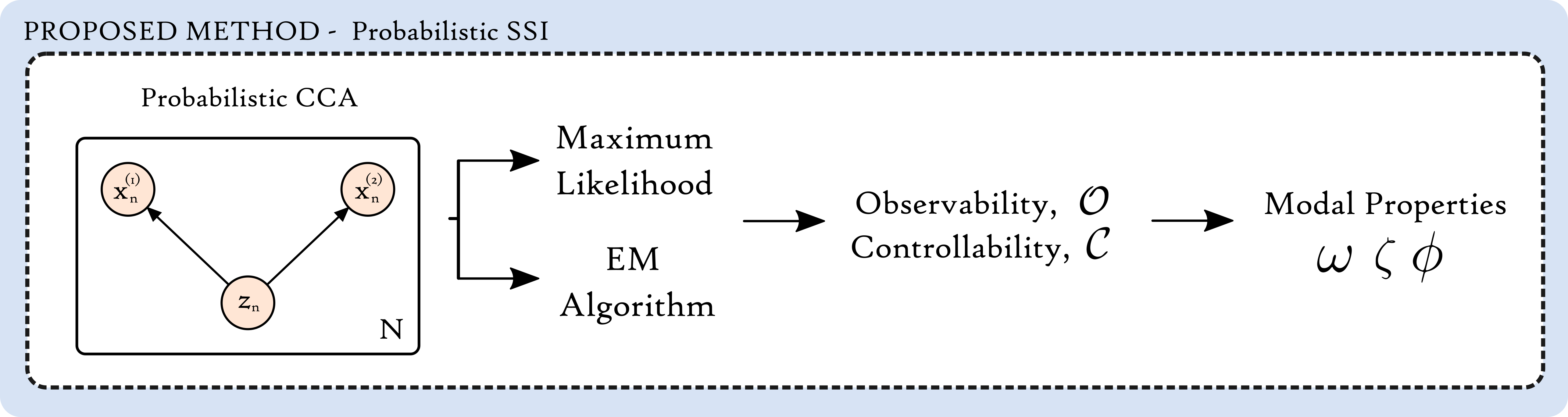}
		\caption{Schematic of the probabilistic SSI procedure, highlighting the replacement of CCA (see Figure \ref{fig:SSI Flowchart}) with probabilistic CCA.}
		\label{fig:Prob-SSI Flowchart}
	\end{figure}

	
Through substitution of CCA with PCCA in the SSI algorithm, where $\mat{X}^{(1)} = \mat{Y}_f$ and $\mat{X}^{(2)} = \mat{Y}_p$, it can be shown that the MLE estimates for the weights $\wghtmle{(m)}$, when written in terms of covariance matrices of the future and past, are (as expected) equivalent to the observability matrix and controllability matrix transposed, respectively (Equations \ref{eq:weight1 probssi} - \ref{eq:weight2 probssi}). The reader is referred back to Equation \ref{eq:ObservReach}.
\begin{align}
	\wghtmle{(1)} \ \ &= \ \ \mat{\Sigma}^{1/2}_{ff}\mat{V}_1\mat{\Lambda}^{1/2}\mat{R} \ \ = \ \ \mathcal{O} \label{eq:weight1 probssi}\\
	\wghtmle{(2)} \ \ &= \ \ \mat{\Sigma}^{1/2}_{pp}\mat{V}_2\mat{\Lambda}^{1/2}\mat{R} \ \  = \ \ \mathcal{C}^{\trans} \label{eq:weight2 probssi}
\end{align}

Having established the relationship between the MLE result for the weight matrices and the observability and controllability matrices, this new probabilistic formulation provides the necessary mathematical foundation upon which new probabilistic and hierarchical SSI algorithms can be implemented. Such algorithms promise to enhance identification performance, aid recovery of uncertainty and/or help extract new unseen information. The first of these new probabilistic approaches to SSI can now be shown, specifically the presentation of a statistically robust method capable of automatically handling outlying data in measured timeseries. 

\section{Robust Probabilistic SSI (Robust Prob-SSI)}\label{sec:RobustSSI}

\subsection{Robust Probabilistic CCA}

\begin{figure}[H]
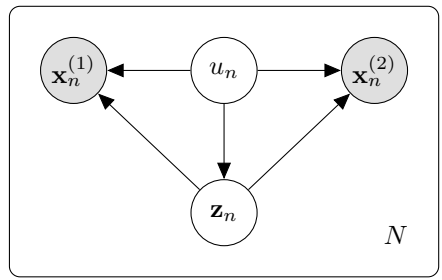

		\centering
	  	\tikz{
			\tikzstyle{latent} = [circle,fill=white,draw=black,inner sep=1pt,
			minimum size=25pt, font=\fontsize{10}{10}\selectfont, node distance=1]
			 \node[latent] (z) {$\vec{z}_n$};%
			  \node[obs,above=of z, xshift=-2cm] (x1) {$\vec{x}_n^{(1)}$}; %
			  \node[obs,above=of z, xshift=2cm] (x2) {$\vec{x}_n^{(2)}$}; %
			  \node[latent,above=of z] (u) {$u_n$}; %
			\tikzstyle{plate caption} = [caption, node distance=0, inner sep=0pt,
				 below left=-8pt and 0pt of #1.south east]
			  \plate [inner sep=0.4cm] {plate1} {(z)(x1)(x2)} {$N$}; %
			  \edge {z} {x1,x2}  
			  \edge {u} {x1}  
			  \edge {u} {x2}  
			  \edge {u} {z}  
	 	} 
	 	\caption{Graphical model for robust probabilistic canonical correlation analysis (robust-PCCA)}
	 	\label{fig:Graph Robust CCA}
	\end{figure}

Standard probabilistic models can be converted to a more `robust' form through the replacement of a Gaussian noise model to that of a Student's t-distribution \cite{Bishop2006}. Building on the probabilistic interpretation of CCA, Archambeau, Delannay and Verleysen \cite{Archambeau2006} constructed a statistically robust equivalent of PCCA. This alternative model is founded upon two key assumptions. The first is that the observed and latent variables can both be modelled by a Student's t-distribution $\mathcal{S}(\vec{\mu},\vec{\Sigma},\nu)$ with mean $\vec{\mu}$ and covariance $\vec{\Sigma}$. The Student's' t-distribution has heavier tails than a typical Gaussian, which are determined by the additional parameter, $\nu$ (degrees of freedom). The presence of heavier tails is preferable as it makes the T-distribution better equipped to handle outliers as it is more likely to capture them within the distribution. The second assumption is that an outlier in the feature space must manifest as an outlier in the latent space. Given this set of assumptions, Archambeau et. al. presented the following probabilistic model (note a continuation of notation from Section \ref{subsec:PCCA})
	\begin{align}	
		\zlatentn{n} \ \ &\sim \ \  \mathcal{S}(\zlatentn{n}|\vec{0},\ident_d,\nu) \\
		\xdatan{n}{(m)}|\zlatentn{n} \ \ &\sim \ \  \mathcal{S}(\xdatan{n}{(m)}|\wght{(m)}\zlatentn{n} + \mean{(m)},\covar{(m)},\nu)
	\end{align}
Making the set of latent variables explicit, the model can be reformulated into the following form
	\begin{subequations}
		\begin{align}	
			u_n \ \ &\sim \ \  \gamdistr{u_n\big|\dfrac{\nu}{2}}{\dfrac{\nu}{2}} \label{eq: robustPCCA distr u} \\
			\zlatentn{n}|u_n \ \ &\sim \ \  \gaussdistr{\zlatentn{n}|\vec{0}}{u_n^{-1}\ident_d} \label{eq: robustPCCA distr z} \\
			\xdatan{n}{(m)}|\zlatentn{n},u_n \ \ &\sim \ \  \gaussdistr{\xdatan{n}{(m)}\big|\wght{(m)}\zlatentn{n} + \mean{(m)}}{u_n^{-1}\covar{(m)}} \label{eq: robustPCCA distr x(m)}\\
			\xdatan{n}{}|\zlatentn{n},u_n \ \ &\sim \ \  \gaussdistr{\xdatan{n}{}|\wght{}\zlatentn{n} + \mean{}}{u_n^{-1}\covar{}}  \label{eq: robustPCCA distr x}
		\end{align}
	\end{subequations}
where $\gamdistr{\alpha}{\beta}$ represents a Gamma distribution, $u_n$ is an additional latent variable and $\wght{}$, $\vec{\mu}$, $\vec{\Sigma}$ and $\nu$ are as previously defined. The corresponding graphical model is shown in Figure \ref{fig:Graph Robust CCA}. 

\subsubsection{Expectation-Maximisation Algorithm}

In contrast to PCCA, direct maximisation of the incomplete data log-likelihood $\sum_n \log p(\vec{x}_n)$ with respect to the parameters is intractable. However, using equations (\ref{eq: robustPCCA distr u}-\ref{eq: robustPCCA distr x}) estimates for the variables can be recovered through an iterative scheme. Archambeau et.al employ the EM approach, maximising the the expected complete-log-likelihood \cite{Archambeau2006}.

\begin{equation}
	\lkhd(\params|\xdatan{n}{},\zlatentn{n},u_n) = \sum^N_{n=1} \ln{p_{}(\xdatan{n}{},\zlatentn{n},u_n | \params)} 
\end{equation}
where $\params = (\mean{},\wght{},\covar{},\nu)$.

The necessary update equations in the E-step (expectation) and M-step (maximisation) of robust PCCA are given by the equations \ref{eq:bar un}-\ref{eq: robustPCCA nu line search}. For completeness, the robust PCCA algorithm has been re-derived\footnote{The reader may notice certain equations differ from those in the original robust projections paper \cite{Archambeau2006}. As is evident from the derivation, there appears to be some typographical errors in the original paper, most notably in the definition of the covariance matrix update equation (Equation 37 in \cite{Archambeau2006})} in \ref{A: Robust PCCA deriv}.  

E-Step
	\begin{align}
		\bar{u}_n \ \ &= \ \ \dfrac{D + \nu}{ u_n(\xdatan{n}{} - \mean{})^{\trans}\mat{A}^{-1}(\xdatan{n}{} - \mean{}) + \nu}
		\label{eq:bar un}\\
		\nonumber\\
		\ln\tilde{u}_n \ \ &= \ \ \psi\left(\dfrac{D + \nu}{2}\right) - \ln\left(\dfrac{u_n(\xdatan{n}{} - \mean{})^{\trans}\mat{A}^{-1}(\xdatan{n}{} - \mean{}) + \nu}{2}\right)\\
		\nonumber\\
		\zbarn \ \ &= \ \ \mat{B}^{-1} \wght{\trans}\covar{-1}(\xdatan{n}{} - \mean{})\\
		\nonumber\\
		\bar{\mat{S}}_n \ \ &= \ \ \mat{B}^{-1} + \bar{u}_n\zbarn\zbarn^{\trans}
	\end{align}
	
	where $\mat{A} = \left(\covar{} + \wght{}\wght{\trans}\right)$, $\mat{B} = \wght{\trans}\covar{-1}\wght{} + \ident_d$, 		$\bar{u}_n = \expect{u_n}$, $\ln\tilde{u}_n = \expect{\ln(u_n)}$, $\bar{\zlatentn{}}_n = \expect{\zlatentn{n}}$, $\bar{\mat{S}}_n = \expect{u_n\zlatentn{n}\zlatentn{n}^{\trans}}$ and $\psi(\cdot)$ denotes the digamma function. 

Subsequently, the update equations for the parameters in the M-step are given by
	\begin{align}
		\begin{split}
			\covar{\prime} \ \ &= \ \ \dfrac{1}{N}\sum^N \biggl[\ubarn (\xdatan{n}{} - \mean{}) (\xdatan{n}{} - \mean{})^{\trans} - \ubarn (\xdatan{n}{} - \mean{})(\wght{}\zbarn)^{\trans} \\ &\qquad\qquad\qquad - \ubarn (\wght{}\zbarn)(\xdatan{n}{} - \mean{})^{\trans} + \ubarn \wght{} \bar{\mat{S}}_n \wght{\trans} \biggr]
		\end{split}\\
		\wght{\prime} \ \ &= \ \ \dfrac{\overset{N}{\sum} \ubarn(\xdatan{n}{} - \mean{})\zbarn^{\trans}}{\overset{N}{\sum} \bar{\mat{S}}_n} 
		\label{eq: robustPCCA W update}\\
		\begin{split}
			\ \mean{\prime} \ \ &= \ \ \dfrac{\overset{N}{\sum} \ubarn (\xdatan{n}{}  - \wght{} \zbarn)}{\overset{N}{\sum} \ubarn}
		\end{split}\\
		\begin{split}
			0 \ \ &= \ \ 1  + \ln\left(\dfrac{\nu}{2}\right) - 2\ \psi\left(\dfrac{\nu}{2}\right) + \dfrac{1}{N}\sum^N \bigl[\ln(\tilde{u}_n) - \ubarn \bigr]
		\end{split}
		\label{eq: robustPCCA nu line search}
	\end{align}

\noindent where $(\cdot)^{\prime}$ denotes an updated parameter and $\nu^{\prime}$ can be found by solving Equation \ref{eq: robustPCCA nu line search} through a suitable line search. The convergence of the EM algorithm was monitored using the Q-function, derived using standard theory \cite{Bishop2006}.

It is important to note that despite the presence of heavier tails, in the case of no outliers, it is expected that the Students' t-distribution will tend towards the PCCA maximum likelihood result. 

\subsection{Robust Probabilistic SSI}

\begin{figure}[h]
	\centering
	\includegraphics[width=0.9\textwidth]{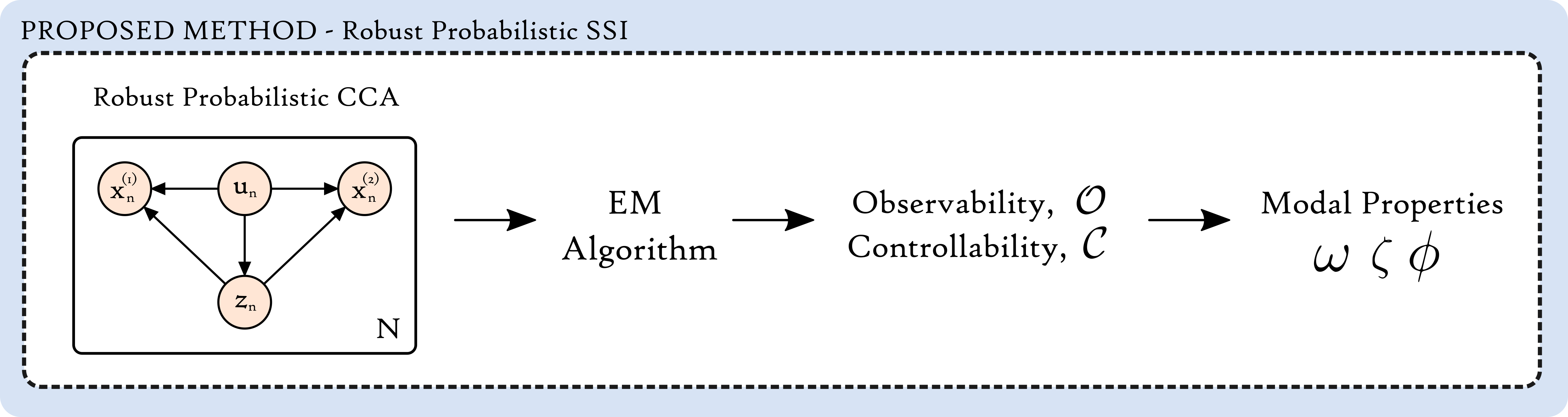}
	\caption{Schematic of the robust probabilistic SSI procedure, highlighting the replacement of PCCA (see Figure \ref{fig:Prob-SSI Flowchart}) with robust PCCA.}
	\label{fig:Robust Prob-SSI Flowchart}
\end{figure}



After successful convergence of the EM algorithm, final estimates for the weights can be obtained. These weights, as described by the model, are considered to be the statistically robust transformations from the latent space to the observation space. The commonality between the weights recovered by PCCA and robust PCCA, based on their ability to transform the data to the relevant subspace, grant a similar replacement of CCA in the SSI procedure, now for robust PCCA.
\begin{align}
	\wght{(1)} \ \ &= \ \ \mathcal{O}\\
	\wght{(2)} \ \ &= \ \ \mathcal{C}^{\trans}
\end{align}
As noted by Archambeau et. al., the weights recovered through EM do not account for the rotational ambiguity seen previously in Section \ref{sec: MLE PCCA}. Although often unnecessary\footnote{This arbitrary rotation is often ignored as its omission still results in data being transformed into the relevant subspace.}, the arbitrary rotation matrix $\mat{R}$ can be recovered through a simple post-processing step, described in \ref{A: Rotation}. In the specific case of robust Prob-SSI, recovery of the rotation matrix is essential for formulating the stabilisation diagram in the usual way. 

\newpage
\section{Results and Discussion}

With a mathematical framework for conducting system identification in a statistically robust and probabilistic way now established, attention can be directed towards robust identification of modal parameters using experimental data. Three separate case studies are used to demonstrate, evaluate and compare the identification performance of robust Prob-SSI to Cov-SSI. The first study (Section \ref{sec: simbenchmark}) uses data from a simulated linear MDOF to benchmark the general performance of robust Prob-SSI to standard Cov-SSI under ideal conditions. The variance exhibited by both methods due to different random input forcing is also briefly explored. The second case study (Section \ref{sec:Corrupted}) exploits the same simulated MDOF system but now containing artificially introduced outliers. This study is used to test the robustness of identification to outliers under known conditions. The third and final case study (Section \ref{sec: z24benchmark}) uses data collected from the Z24 bridge benchmark. This study is used to investigate the method's overall appropriateness to ``real-world'' data and assess and evaluate its overall performance against Cov-SSI.

In all tests, robust Prob-SSI was directly compared to Cov-SSI to highlight differences in identification performance. The reader will notice the omission of results from Prob-SSI. As has been shown, the MLE estimates obtained through Prob-SSI are equivalent to those found through Cov-SSI and so their addition was redundant. 

Computational efficiency was not a primary factor of this work as it was deemed to be a distraction from the hypothesis being proposed. However, due to the iterative nature of the EM algorithm and the variability in convergence, robust Prob-SSI at present displays typically longer computation times, when compared to the SVD based Cov-SSI. Nevertheless, through effective computational optimisation, it is believed this could be greatly improved. 

\subsection{Benchmark: Simulated MDOF Linear System} \label{sec: simbenchmark}

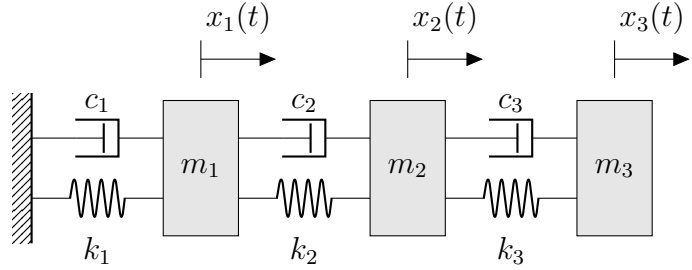
\begin{figure}[H]
\begin{center}
\begin{circuitikz}
	\pattern[pattern=north east lines] (0,0) rectangle (0.25,2);
	\draw[thick] (0.25,0) -- (0.25,2);
	
	\draw(0.25,0.6) to[spring, l_=$k_1$] (2,0.6);
	\draw(0.25,1.4) to[damper, l=$c_1$] (2,1.4); 
	\draw[fill=gray!20] (2,0.1) rectangle (3,1.9);
	\node at (2.5,1) {$m_1$};
	\draw (2.5,2.2) -- (2.5,2.7);
	\draw[->] (2.5,2.45) -- (3.5,2.45);
	\node at (3,3) {$x_1(t)$};

	\draw(3,0.6) to[spring, l_=$k_2$] (4.75,0.6);
	\draw(3,1.4) to[damper, l=$c_2$] (4.75,1.4); 
	\draw[fill=gray!20] (4.75,0.1) rectangle (5.75,1.9);
	\node at (5.25,1) {$m_2$};
	\draw (5.25,2.2) -- (5.25,2.7);
	\draw[->] (5.25,2.45) -- (6.25,2.45);
	\node at (5.75,3) {$x_2(t)$};

	\draw(5.75,0.6) to[spring, l_=$k_3$] (7.5,0.6);
	\draw(5.75,1.4) to[damper, l=$c_3$] (7.5,1.4); 
	\draw[fill=gray!20] (7.5,0.1) rectangle (8.5,1.9);
	\node at (8,1) {$m_3$};
	\draw (8,2.2) -- (8,2.7);
	\draw[->] (8,2.45) -- (9,2.45);
	\node at (8.5,3) {$x_3(t)$};
\end{circuitikz}
\end{center}
\caption{Simulated MDOF System}
\label{fig:Simulated MDOF System}
\end{figure}

To evaluate the identification performance of robust Prob-SSI, a suitable benchmark was created. Response data was generated using a generic three degree-of-freedom (DOF) linear dynamic system with proportional damping; shown in Figure \ref{fig:Simulated MDOF System} and described by the model parameters in Equation \eqref{eq: 3DOF}. The system was subsequently excited using broadband forcing, of order $10^{-2}$ with a standard deviation of 1, mimicking ambient excitation. The system was simulated at a sample rate of $1 \times 10^3$ Hz and generated 8192 samples. The Hankel matrix was constructed using 20 lags; i.e. 10 forwards and 10 backwards in time.

\begin{eqnarray}
		\mat{M} = \begin{bmatrix} m & 0 & 0\\ 0 & m & 0 \\ 0 & 0 & m \end{bmatrix}\ , \ \ 
		\mat{K} = \begin{bmatrix} 2k & -k & 0\\ -k & 4k & -0.5k \\ 0 & -0.5k & k \end{bmatrix} \ , \ \ 
		\mat{C} = \mat{K} \times 10^{-4}
		\label{eq: 3DOF}
\end{eqnarray}
\begin{eqnarray*}
	k = 1\times 10^4 \ \mathrm{(N/m)}, \ \ m = 10 \ \mathrm{(kg)}
\end{eqnarray*} 

Before application of robust Prob-SSI, suitable initial conditions are required for the EM algorithm. A small perturbation to the CCA (PCCA maximum likelihood) result was chosen as the initial estimate of the full weight matrix, providing a sensible prior that, one would expect, converges to a solution sooner and therefore reduces computational expense. The initial estimates for the covariance matrices were randomly sampled from an inverse Wishart distribution $\mathcal{W}^{-1}(\mat{K},\nu)$ where $\nu = D + 2$ and $\mat{K} = \ident_D$.

\begin{figure}[H]
	\centering
	\includegraphics[width=0.9\textwidth]{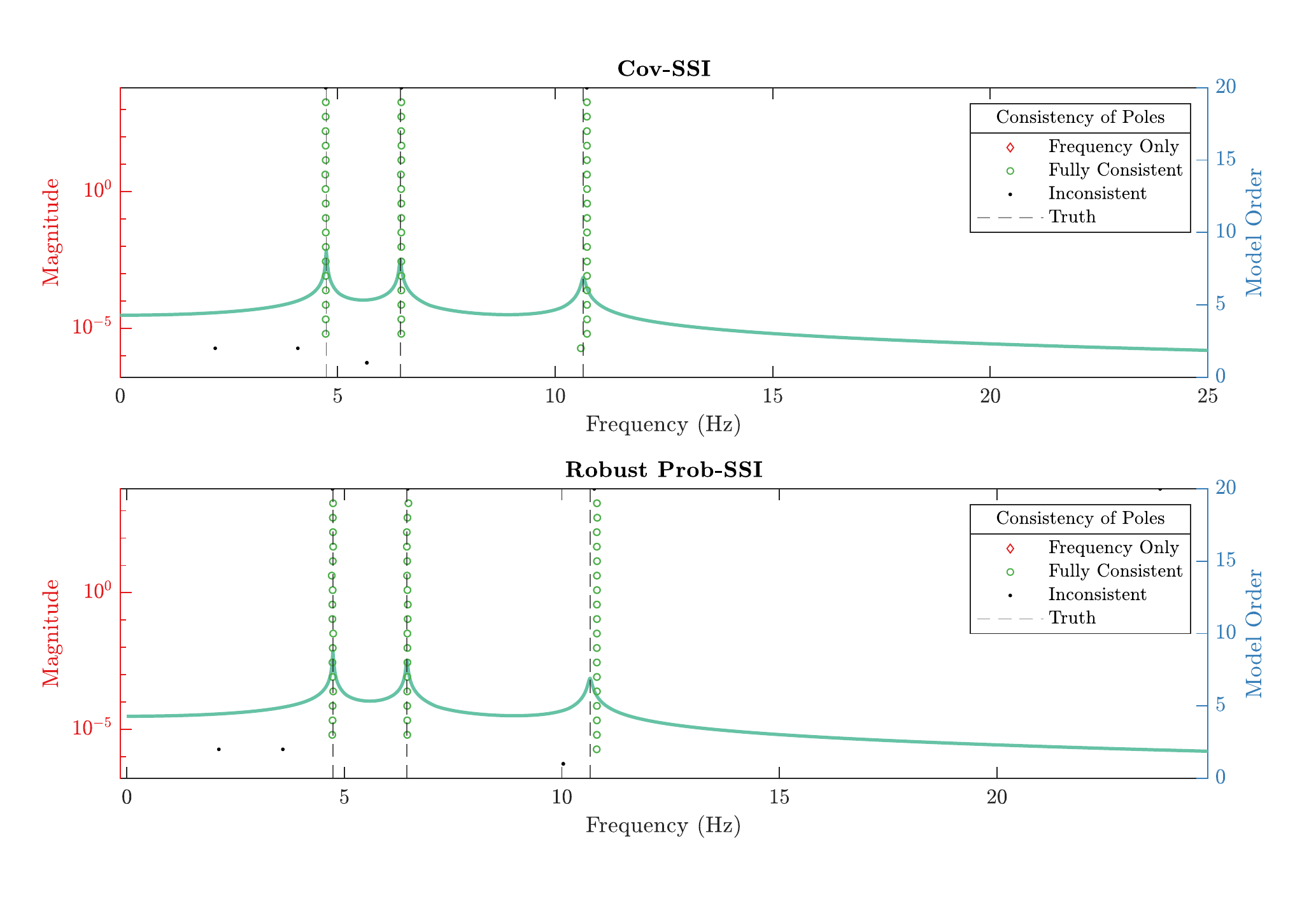}
	\caption{Consistency diagrams for the 3DOF system, recovered using Cov-SSI (top) and robust Prob-SSI (bottom).}
	\label{fig:StabilDiag Benchmark}
\end{figure}

Following analysis of the data, consistency diagrams\footnote{This often referred to as the stabilisation diagram, however the authors believe this choice of terminology may cause confusion with the `stability' of poles referenced in control theory. Hence, the term consistency diagram is used. This transition has also been adopted by others in the field of dynamics \cite{allemang2010}} were generated for both methods using standard techniques. These are shown in Figure \ref{fig:StabilDiag Benchmark}. To ensure fair comparison, fixed definitions for the consistency criteria were applied to both methods. This ensured that any discrepancies in identified consistency arose solely from the method and not through independent changes to specific criteria. The chosen criteria were: a $2\%$ relative change in frequency, a $5\%$ absolute change in damping ratio, and a $98\%$ relative correspondence in the MAC value. 

As evident from Figure \ref{fig:StabilDiag Benchmark}, the found consistency diagram for Robust Prob-SSI is very comparable with that of Cov-SSI. Clear columns of fully consistent poles are easily identifiable and, as shown in Table \ref{tab:Freqs and Damp}, the estimated values for the natural frequencies and damping ratios comfortably lie within an acceptable tolerance of the ground truth used in the simulation.

\newcolumntype{C}[1]{>{\centering\let\newline\\\arraybackslash\hspace{0pt}}m{#1}}
\begin{table}[h]
	\caption{A comparison of the natural frequencies and damping ratios obtained in the benchmarking study.}
	\vspace{0.25cm}
	\centering
		\begin{tabular}{@{} r C{1.5cm} C{1.5cm} C{1.5cm} c C{1.5cm} C{1.5cm} C{1.5cm}@{}}\toprule
			& \multicolumn{3}{c}{Natural Frequency (Hz)} & & \multicolumn{3}{c}{Damping Ratio} \\ \cmidrule(r){2-8}
			& 1 & 2 & 3 && 1 & 2 & 3 \\ \midrule
			Truth & 4.74 & 6.44 & 10.65 && 0.0033 & 0.0020 & 0.0015\\
			Cov-SSI & 4.73 & 6.46 & 10.73 && 0.0024 & 0.0050 & 0.0128\\
			Robust Prob-SSI & 4.74 & 6.45 & 10.80 && 0.0030 & 0.0047 & 0.0100\\ \bottomrule
	\end{tabular}
	\label{tab:Freqs and Damp}
\end{table}

\begin{figure}[h]
	\centering
	\includegraphics[width=\textwidth]{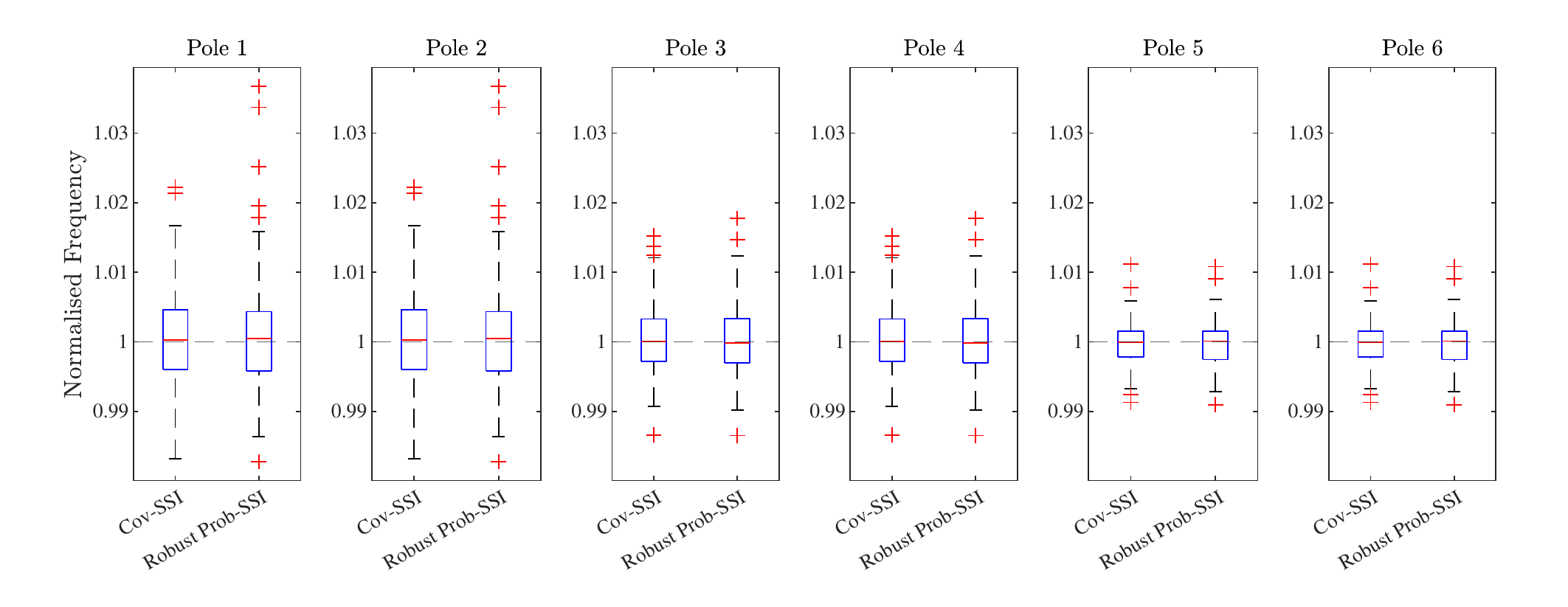}
	\caption{A demonstration of the variance seen in the natural frequency estimates using Cov-SSI and robust Prob-SSI for 100 tests on the same 3DOF dynamic system but with varying input forcing.}
	\label{fig:Var in Nat Freq}
\end{figure}

Due to stochastic variation in the excitation and, in the case of Robust Prob-SSI, the initialisation of the EM algorithm, the inherent variance shown by both methods was also explored. Assuming a correct number of DOF, both methods were applied to 100 datasets whereby the underlying system dynamics remain unchanged but the random seed needed to generate the random forcing varied between tests. The estimates of the natural frequencies from these tests are shown in Figure \ref{fig:Var in Nat Freq}.

As expected, due to the complex nature of the system poles, the recovered estimates exist in conjugate pairs. This is evident from the clear matching of results seen in both tests between poles 1 and 2, poles 3 and 4, and poles 5 and 6. The variance exhibited by robust Prob-SSI is highly similar to that of Cov-SSI for all the recovered poles, when the correct number of degrees of freedom is assumed. Overall, this result demonstrates that robust Prob-SSI is capable of accurately replicating the results of Cov-SSI. 

\subsection{Corrupted Dataset: Simulated MDOF Linear System} \label{sec:Corrupted}

Having established that robust Prob-SSI is capable of replicating the results of Cov-SSI for a simulated, `clean' case, we now turn our attention to the proposed robustness of this method to corrupted datasets. In the context of this work, a `corrupted' dataset will refer to any dataset containing either artificially generated or naturally occurring, atypical observations (outliers\footnote{An important detail worth noting is the definition of outliers. Archambeau et.al. assume that an outlier in the latent space will manifest as an outlier in the dataspace. This is a relatively sensible assumption given the linear transformation to the data. However, recall that Cov-SSI relies on components of the cross-covariance matrix. Therefore, if the cross-covariance is insensitive to atypical observations present in the timeseries, there will be no discernible effect on the identification procedure. In the simulated case, this can be explored, but for real datasets, the `clean' case is unobtainable.}). Furthermore, the terms `atypical observations' and `outliers' will be used interchangeably. 

The corrupted dataset used in this study is based on the same linear dynamic MDOF system used in the benchmarking case but now with the inclusion of artificial outliers of a chosen type and dispersion. In the example shown here, outliers were introduced at random locations in each sensor channel and set to a specified value, plus some small amount of noise. This was intended to mimic random sensor dropout where the measured value is pinned to the lower supply rail of the DAQ unit with some noise. As one might expect, there are many different types and patterns of outliers that could manifest in experimental data and it would be extremely difficult to generate and analyse all such cases. Nevertheless, for completeness, a selection of corrupted datasets with varying outlier types were also generated and analysed. The results of these series of tests are presented in \ref{A: Varying Outlier Study}.

\begin{figure}[h]
	\centering
	\includegraphics[width=\textwidth]{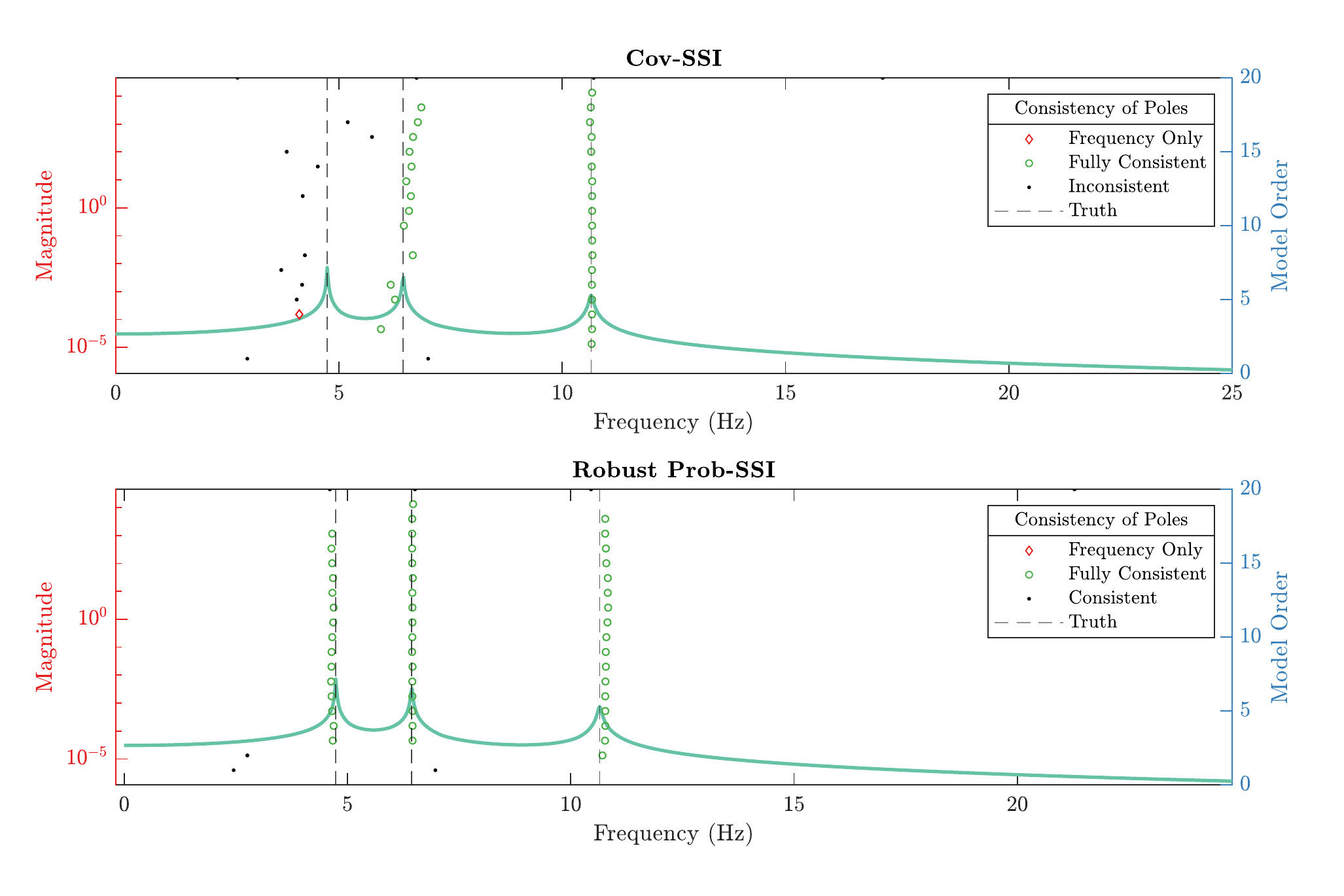}
	\caption{Consistency diagrams recovered using Cov-SSI (top) and robust Prob-SSI (bottom) using response data from a simulated 3DOF `corrupted' dataset, containing 0.1\% artificially introduced, randomly located outliers (in each channel), set to a specified value, plus some small amount of noise.}
	\label{fig:Stabil Diagrams Outliers}
\end{figure}

In the case shown, the number of outliers was set to 0.1\% per channel which, for a dataset with 3 sensor channels and 8192 data points, equated to approximately 24 outliers in total. The response data and outliers are shown in Figure \ref{fig: response random}. To ensure a fair comparison of the two approaches, definitions for the consistency criteria were once again fixed. The consistency diagrams recovered from this scenario are shown in Figure \ref{fig:Stabil Diagrams Outliers}. 

\begin{figure}[H]
	\centering
	\includegraphics[width=0.9\textwidth]{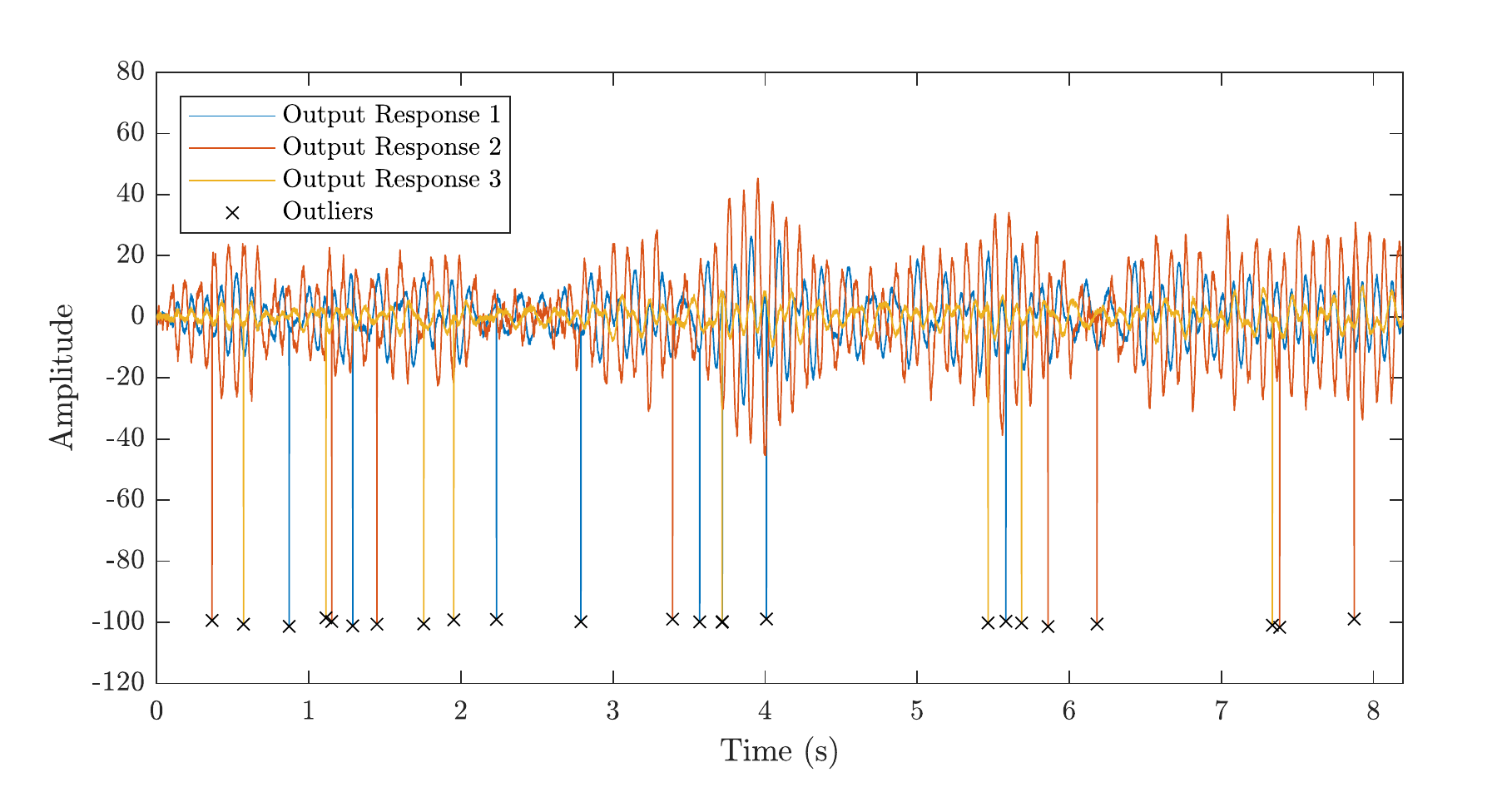}
	\caption{Response data from a simulated 3DOF `corrupted' dataset, containing 0.1\% artificially introduced, random outliers in each signal channel. The value of the outliers was set to a specified value plus some small amount of noise, intended to mimic random sensor dropout where the measured value is pinned to the lower supply rail of the DAQ unit with some noise}
	\label{fig:	response random}
\end{figure}

It is immediately apparent that robust Prob-SSI displays a significant improvement in identification performance over Cov-SSI when confronted with the corrupted dataset. Whilst robust Prob-SSI comfortably identifies all three correct natural frequencies of the system, traditional Cov-SSI fails to find the first, struggles to accurately identify the second and only successfully identifies the third natural frequency. This failure in identification can be clearly observed as the lack of a column of fully consistent poles in the upper frame of Figure \ref{fig:Stabil Diagrams Outliers}.

However, it is expected that there will be several instances where, despite the presence of outliers, SSI will continue to perform as expected. Similarly, there will also be cases where robust Prob-SSI will also fail to identify the system. The replacement of SSI with robust Prob-SSI is not a straightforward one. In \ref{A: Pct Outlier Study}, it is illustrated how increasing the percentage of outliers inevitably leads to a deterioration in performance for both methods. Clearly, although robust Prob-SSI offers some protection, in the presence of many outliers it will be impossible to identify the dynamics.

\begin{figure}[h]
	\centering
	\includegraphics[width=\textwidth]{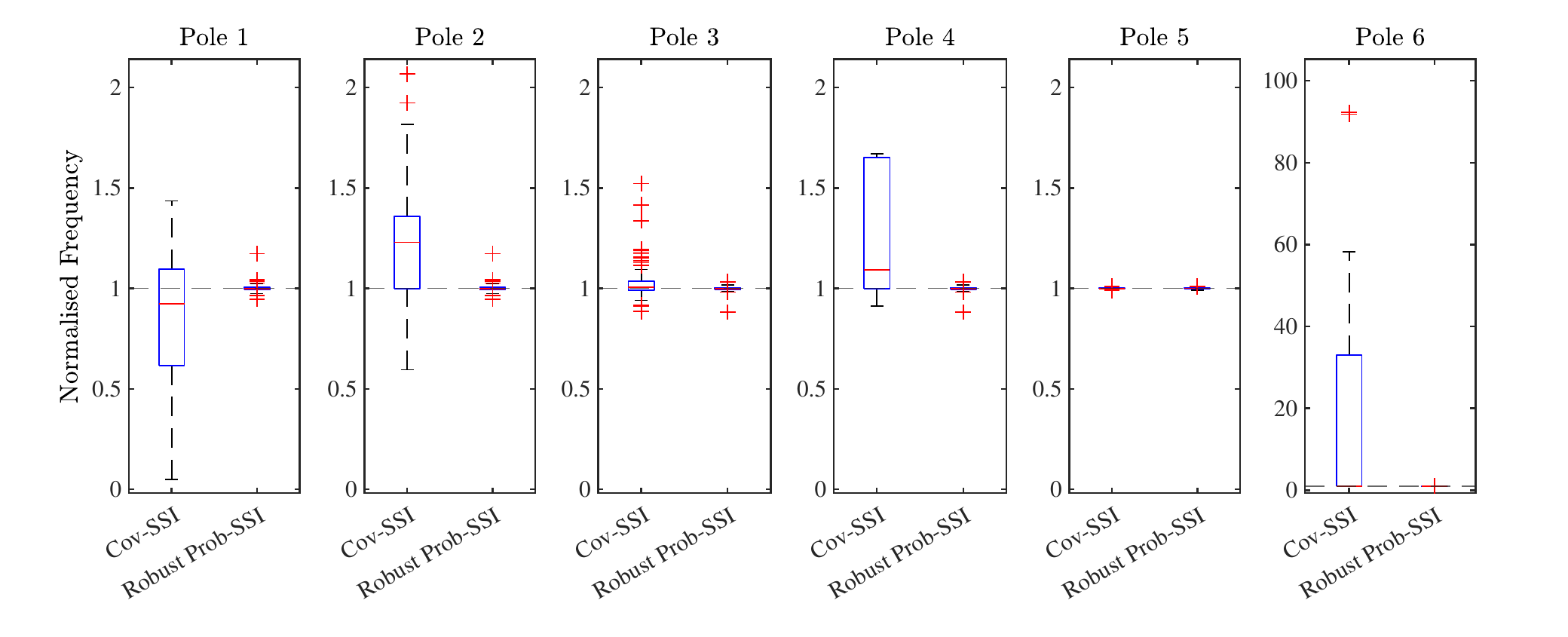}
	\caption{A demonstration of the variance seen in the natural frequency estimates when using Cov-SSI and robust Prob-SSI to analyse response data from 100 datasets of the same 3DOF dynamic system with varying input forcing but with artificially induced random outliers (0.1\% per channel).}
	\label{fig:Var in Nat Freq with Outliers}
\end{figure}

Similar to the benchmark case, the variance of the estimates from both tests was also explored and similarly the correct number of DOF was assumed. The variance in the resulting natural frequency estimates is shown in Figure \ref{fig:Var in Nat Freq with Outliers}. As is clearly shown, the variance in Cov-SSI is significantly greater than that of robust Prob-SSI, 

\newpage
\subsection{Case Study: Z24 Bridge} \label{sec: z24benchmark}

This final study is used to investigate and evaluate the performance of robust Prob-SSI when confronted with a real world example; a subset of the ever dependable Z24 bridge dataset. More information on the Z24 bridge, the available datasets and references to the original work, can be found here (bwk.kuleuven.be/bwm/z24/z24) \cite{Maeck2003,Maeck2001b,Peeters2001a}. Data collected from the Z24 bridge is often employed for new SHM tasks concerned with temporal changes to the modal properties, induced by damage and/or temperature. Its frequent application across structural dynamic research for benchmarking new system identification approaches to large scale structures, in SHM and automatic pole selection, make it a sensible choice for this study.

\begin{figure}[H]
	\centering
	\includegraphics[width=0.8\textwidth]{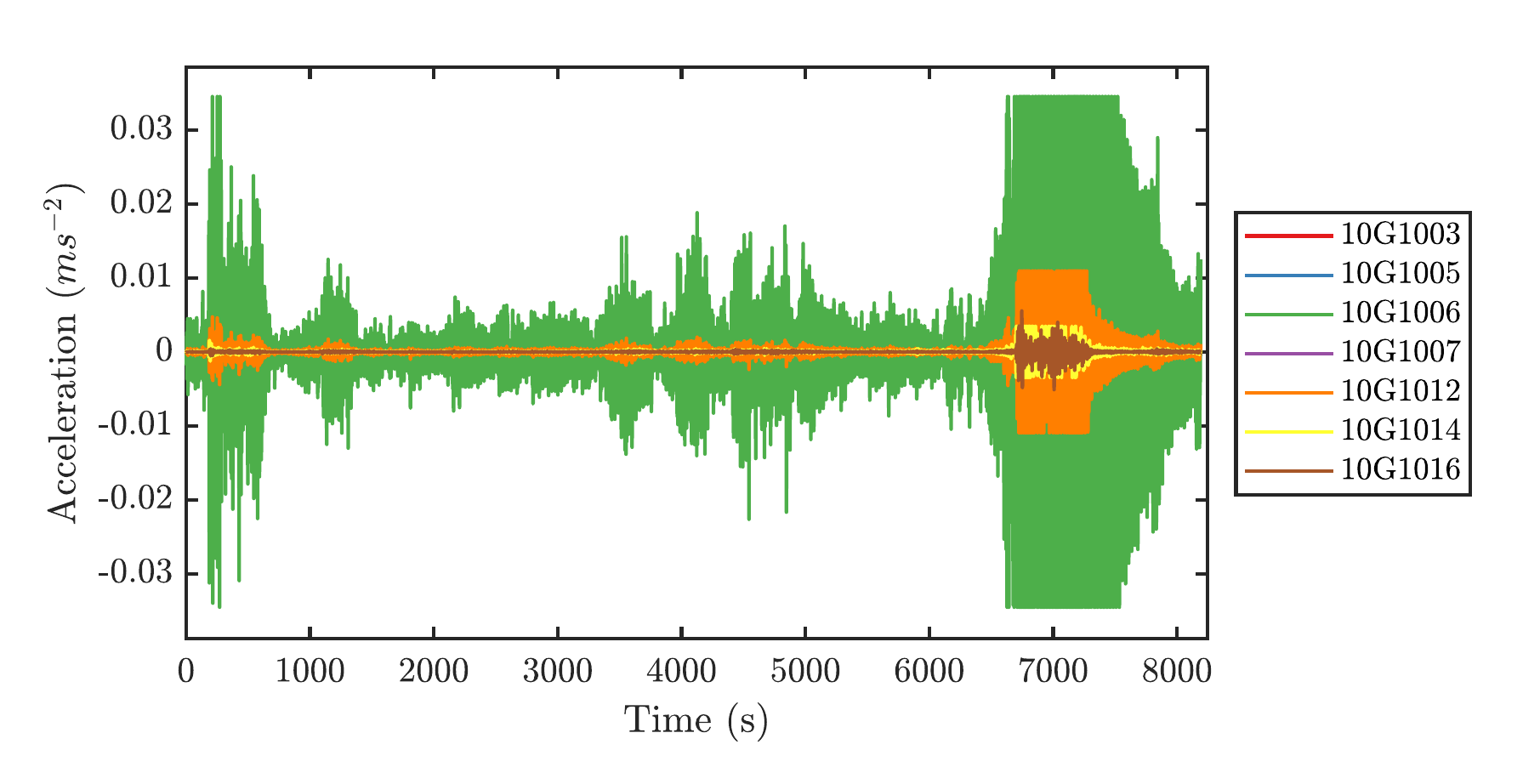}
	\caption{A sample of output time series acceleration measurements (the first 8192 points) taken from the Z24 bridge dataset 10G10 (corresponding to data recorded on `17-Jan-1998' at `10:00:00'), that appear to demonstrate over-ranging of the sensors and therefore clipping of the measured signals.}
	\label{fig:Z24 Accel Measurements}
\end{figure}

It is unknown if the Z24 data features any outlying observations or signal corruption. However, multiple output acceleration signals, across multiple tests in the initial stages of testing, appear to demonstrate clipping (see Figure \ref{fig:Z24 Accel Measurements}). It is undetermined in previous literature if this clipping affects the identification and fundamentally this will remain unknown. However, this interesting feature of the Z24 data makes it highly suitable for our investigation into robust Prob-SSI as a robust alternative to SSI given possible a-typical observations.

\begin{figure}[h]
	\centering
	\includegraphics[width=\textwidth]{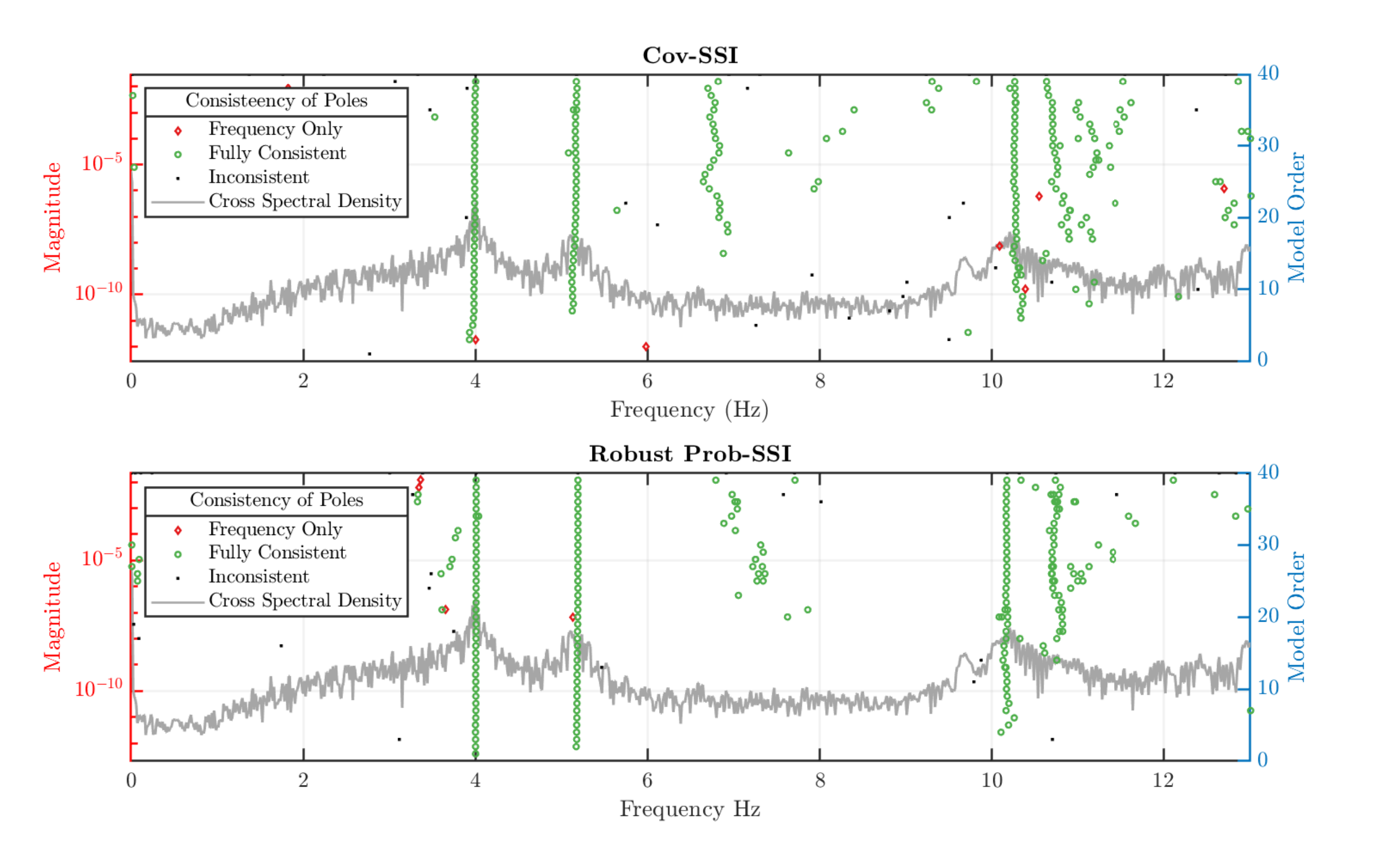}
	\caption{Consistency diagrams generated using results from Cov-SSI (top) and robust Prob-SSI (bottom) after application to a single subset of the Z24 bridge dataset (07E01); corresponding to data recorded on `06-Dec-1997' at `01:00:00'}
	\label{fig: Z24 single stabildiag}
\end{figure}

A subset of the Z24 dataset, specifically data folder 07E01 containing acceleration measurements observed on the `06-Dec-1997' at `10:00:00', prior to the induced damage, was initially selected at random for processing. Only the initial segment, the first 8192 points, was used in the analysis. The resulting consistency diagrams generated using both SSI methods are shown in Figure \ref{fig: Z24 single stabildiag}. The frequency range is limited to show the first 4 natural frequencies, as identified in the original works, as these are often the main frequencies used when analysing this dataset. 

The first two identified modes are easily observable in both diagrams, given the presence of highly consistent poles. The same can also be said of the third and fourth modes, with robust Prob-SSI demonstrating better consistency when compared to Cov-SSI over increasing model order. Across all the identified modes, one could argue that robust Prob-SSI identifies consistent poles at all of these modes at lower model orders than Cov-SSI and thus, would provide more confidence to the practitioner. There is some speculation in the SHM community regarding the existence of a mode at $~7$Hz, however, it is believed to be illusive. Interestingly, the robust approach appears to largely ignore the presence of this mode suggesting it may be the result of noise, aptly accounted for in this case by the Student's t noise model.

This promising initial evidence suggests that robust Prob-SSI exhibits comparatively better identification performance over Cov-SSI. However, a single dataset alone is insufficient to make such a claim. To better assess the overall identification performance to a larger range of cases, Cov-SSI and robust Prob-SSI were applied to the first segment (first 8192 points) of multiple Z24 datafiles, specifically those dated 21st November 1997 14:00:00 through to 25th April 1998 14:00:00, and the resultant poles extracted.

Rather than manually selecting consistent poles in each consistency diagram, a very time consuming activity, automatic selection of the consistent poles was achieved using an implementation of the clustering algorithm DBSCAN. The application of DBSCAN (or derivative thereof, OPTICS) to pole selection in stabilisation diagrams is not a novel one, existing work can be found here \cite{Boroschek2019}. As the performance of any given clustering technique was not the focus of this study, the decision to use DBSCAN clustering over another was arbitrary, however as a very common mode of clustering and one frequently cited in literature, it may seem the most sensible. For information on the theory and implementation of DBSCAN, the reader is directed here \cite{ester1996}. 

Akin to the approach used by Boroschek and Bilbao \cite{Boroschek2019}, the distance metric in DBSCAN was chosen to take the following form:

\begin{equation}
	dist(p_i,p_j) = \dfrac{\abs{\omega_i - \omega_j}}{\mathrm{max}(\omega_i,\omega_j)} + (1 - \mathrm{MAC}(\phi_i,\phi_j))
\end{equation}

\noindent where the natural frequencies $\omega_i, \omega_j$ and mode shapes $\phi_i, \phi_j$ correspond to poles $p_i$ and $p_j$ respectively, and the MAC is the modal assurance criterion. The reachability distance was chosen to be $0.005\%$, with the minimum number of objects set to 25.  

\begin{figure}[h]
	\centering
	\includegraphics[width=\textwidth]{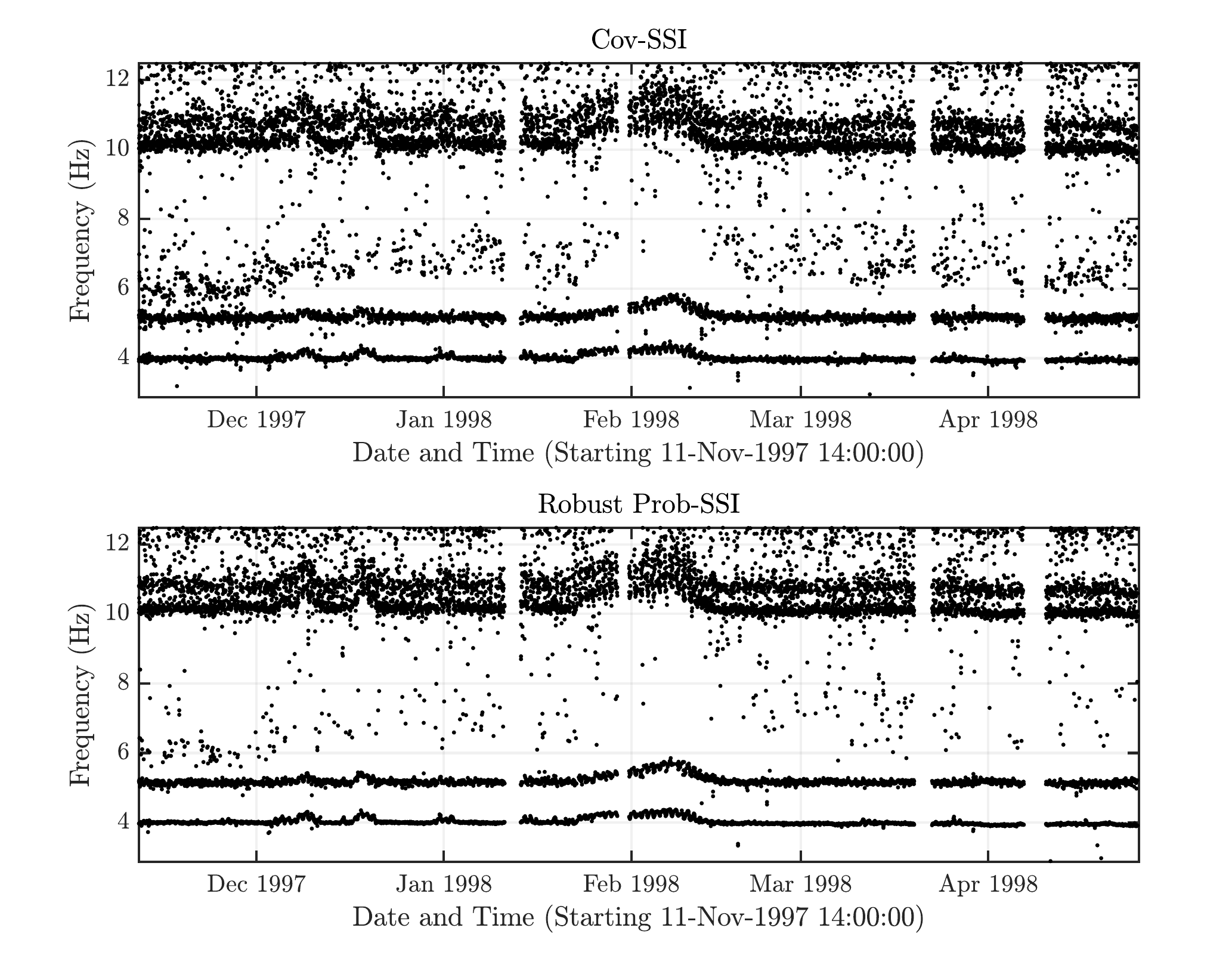}
	\caption{Temporal changes to the natural frequency estimates of the Z24 bridge, identified using Cov-SSI (top) and Robust Prob-SSI (below). DBSCAN clustering was used to extract the non-spurious, consistent poles from each individual dataset. Both methods were applied to the first segment (first 8192 points) of all datafiles ranging from 21st November 1997 14:00:00 to 25th April 1998 14:00:00.}
	\label{fig: Z24 stabildiag}
\end{figure}

Following application of DBSCAN to the Z24 data, centres of the recovered pole clusters were plotted against the corresponding date and time of collection. Figure \ref{fig: Z24 stabildiag} displays the temporal changes to the natural frequency estimates obtained using Cov-SSI and robust Prob-SSI. The first noticeable difference between the two plots in Figure \ref{fig: Z24 stabildiag} is variation in the number of consistent poles situated around 7 Hz, corresponding to the supposedly illusive mode. Congruent to the assessment of the single consistency diagram, robust Prob-SSI (bottom) appears to find a lower number of consistent poles at this frequency across the full range of tests, when compared to Cov-SSI (top). Again, this is likely due to any noise generating this nuisance mode, being accounted for by the Student's t noise model of robust Prob-SSI.

Turning attention to the first two modes, much lower variance is observed in the frequency estimates recovered using robust Prob-SSI across the entire test series compared to Cov-SSI. This reinforces preliminary conclusions drawn from Figure \ref{fig: Z24 single stabildiag} regarding the increased consistency of recovered poles using robust Prob-SSI. This reduction in variance is also apparent, albeit less obviously, in modes three and four, suggesting robust Prob-SSI is capable of recovering less noisy pole estimates than Cov-SSI. This has several benefits, not least in providing the experienced practioner with more confidence when selecting suitable poles from the consistency diagram. This new technique may also prove to be a preferred method of modal parameter recovery for SHM schemes tasked with accurately monitoring temporal changes to modal properties and inferring possible damage. Reducing variance in modal property estimates is of value, in the SHM setting, as it offers a route to reducing false positives and increasing sensitivity in anomaly detection.




\section{Conclusions}

This paper presented a new formulation of the stochastic subspace identification (SSI) algorithm, recasting it as a problem in probabilistic inference. Such an approach was made possible through close alignment of SSI with the theory of probabilistic projections \cite{Bach2005}. Mathematically, an equivalence between the maximum likelihood estimates of the weights (linear transformations) of the latent projections, and the observability and controllability matrix transposed, was established. This unique perspective of SSI lays the requisite mathematical foundation required to enable a multitude of new SSI-based OMA algorithms, likely to be constructed using more complex probabilistic and hierarchical techniques. Such algorithms could add more value to the inclusion of SSI as a component in broader probabilistic frameworks, although the detail of which not discussed here. 

Based on this new probabilistic form, this paper then introduced a second contribution - a statistically robust SSI algorithm, capable of providing a principled and automatic way of handling atypical observations in multi-output time-series responses, such as those observed in field measurements. Consequently, the proposed method was evaluated against three primary case studies. The first study benchmarked robust Prob-SSI against conventional SSI using a familiar MDOF dataset and displayed highly comparable results in the case where data are not corrupted. The second study sought to evaluate the methods ability to resist misidentification when presented with outliers. Robust Prob-SSI was exposed to a 'corrupted' dataset containing artificially introduced outliers; designed to mimic a typical form of anomalous data observed in experimental testing. Subsequent analysis showed robust Prob-SSI outperform conventional SSI, with the former exhibiting better modal identification performance (consistency) and higher confidence in the found poles. The last study investigated data taken from the Z24 Bridge benchmark, chosen with the view of better demonstrating identification performance on data more indicative of a real system. Results from this analysis showed that robust Prob-SSI, enabled by the novel probabilistic interpretation of the SSI algorithm, gives an automated and consistent approach to managing data collected in OMA settings.

This initial body work highlighted the first of many new formulations of SSI, made possible through the novel viewpoint of SSI as a problem in probabilistic inference. The authors envisage future work pursuing new algorithms, constructed upon Prob-SSI, for example, adopting a Bayesian methodology through the inclusion of priors over the model parameters or alternative hierarchical structures.

\section*{CRediT Authorship Contribution Statement}

\textbf{Brandon O'Connell:} Conceptualization, Methodology, Software, Validation, Visualization, Writing - Original Draft \textbf{Timothy Rogers:} Conceptualization, Software, Writing - Review \& Editing, Supervision, Funding acquisition

\section*{Declaration of Competing Interest}

The authors declare that they have no known competing financial interests or personal relationships that could have appeared to influence the work reported in this paper.

\section*{Acknowledgements}

The authors gratefully acknowledge the support of the Engineering and Physical Sciences Research Council (EPSRC), UK through grant number EP/W002140/1. The authors also thank Elizabeth Cross and Ulf Tygesen for their insight and many helpful discussions. For the purpose of open access, the authors have applied a Creative Commons Attribution (CC BY) license to any Author Accepted Manuscript version arising.

\newpage
\appendix

\section{Robust Probabilistic CCA Derivation}\label{A: Robust PCCA deriv}

\begin{equation}
	\begin{split}
		\left < \mathcal{L}\right > = \sum^N \log{p(x_n,z_n)}
	\end{split}
\end{equation}

\begin{figure}[h]
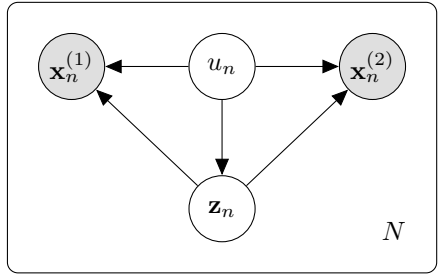

	\centering
	  \tikz{
		\tikzstyle{latent} = [circle,fill=white,draw=black,inner sep=1pt,
		minimum size=25pt, font=\fontsize{10}{10}\selectfont, node distance=1]
		 \node[latent] (z) {$\zlatentn{n}$};%
		  \node[obs,above=of z, xshift=-2cm] (x1) {$\xdatan{n}{(1)}$}; %
		  \node[obs,above=of z, xshift=2cm] (x2) {$\xdatan{n}{(2)}$}; %
		  \node[latent,above=of z] (u) {$u_n$}; %
		\tikzstyle{plate caption} = [caption, node distance=0, inner sep=0pt,
			 below left=-8pt and 0pt of #1.south east]
		  \plate [inner sep=0.4cm] {plate1} {(z)(x1)(x2)} {$N$}; %
		  \edge {z} {x1,x2}  
		  \edge {u} {x1}  
		  \edge {u} {x2}  
		  \edge {u} {z}  
	 } 
	 \caption{Graphical latent model for Robust Canonical Correlation Analysis}
\end{figure}

\begin{subequations}
	\begin{align}	
		p(u_n) &= \gamdistr{u_n\big|\dfrac{\nu}{2}}{\dfrac{\nu}{2}}\\
		p(\zlatentn{n}|u_n) &= \gaussdistr{\zlatentn{n}|\vec{0}}{u_n^{-1}\ident_d}\\
		p(\xdatan{n}{(m)}|\zlatentn{n},u_n) &= \gaussdistr{\xdatan{n}{(m)}\big|\wght{(m)}\zlatentn{n} + \mean{(m)}}{u_n^{-1}\covar{(m)}}\\
		p(\xdatan{n}{}|\zlatentn{n},u_n) &= \gaussdistr{\xdatan{n}{}|\wght{}\zlatentn{n} + \mean{}}{u_n^{-1}\covar{}}
	\end{align}
\end{subequations}

where $\xdatan{n}{} = [\xdatan{n}{(1)};\xdatan{n}{(2)}]$, $\xdata{(m)} = [\xdatan{1}{(m)}, \dots,\xdatan{N}{(m)}]$ with $m = 1,2$, $\xdata{} = [\xdata{(1)};\xdata{(2)}]$, $\zlatent{} = [\zlatentn{1},\dots.\zlatentn{N}]$ and $\covar{}$ is a block-diagonal covariance matrix with matrices $\covar{(1)}$ and $\covar{(2)}$ as the diagonal elements.\\

In the original Robust Projections paper, the normal distributions are defined using the precision matrix, rather than covariance matrix. Note that here only covariance matrices are used. 

To derive the EM update equations for the parameters and an appropriate Q-function to monitor convergence of the EM algorithm, the log-likelihood function must be defined:

\begin{equation}
	\lkhd(\params|\xdatan{n}{},\zlatentn{n},u_n) = \sum^N \ln{p_{}(\xdatan{n}{},\zlatentn{n},u_n | \params)} 
\end{equation}

where $\params = (\mean{},\wght{},\covar{},\nu)$.





\begin{equation}
	\begin{split}
		\lkhd(\params) = & \ \dfrac{1}{N}\sum^N \biggl[ -\dfrac{D}{2}\ln{(2\pi)} - \hlf\ln{\abs{u_n^{-1}\ident_d}} - \hlf(\zlatentn{n} - \vec{0})^{\trans}(u_n^{-1}\ident_d)^{-1}(\zlatentn{n} - \vec{0}) \\ & \dots  + \dfrac{\nu}{2}\ln\left(\dfrac{\nu}{2}\right) - \ln\left(\Gamma\left(\dfrac{\nu}{2}\right)\right) + \dfrac{\nu}{2}\ln(u_n) - \dfrac{\nu}{2}u_n - \dfrac{D}{2}\ln{(2\pi)}  \\ & \dots - \hlf\ln{\abs{u_n^{-1}\covar{}}} - \hlf\bigl(\xdatan{n}{} - (\wght{}\zlatentn{n} + \mean{})\bigr)^{\trans}(\inv{u_n}\covar{})^{-1}\bigl(\xdatan{n}{} - (\wght{}\zlatentn{n} + \mean{})\bigr) \biggr]
	\end{split}
	\label{eq: expanded loglkhd1}
\end{equation}

Through the simplification of terms, Eq.\ref{eq: expanded loglkhd1} can be rearranged and simplified to

\begin{equation}
	\begin{split}
		\lkhd(\params) = & \ \dfrac{1}{N}\sum^N \biggl[ -D\ln{(2\pi)} - \hlf\ln{\abs{u_n^{-1}\ident_d}} - \hlf u_n \zlatentn{n}^{\trans}\zlatentn{n} + \dfrac{\nu}{2}\ln\left(\dfrac{\nu}{2}\right) \\ & \hdots - \ln\left(\Gamma\left(\dfrac{\nu}{2}\right)\right) + \dfrac{\nu}{2}\ln(u_n) - \dfrac{\nu}{2}u_n - \hlf\ln{\abs{u_n^{-1}\covar{}}}  \\ & \hdots - \hlf u_n \bigl((\xdatan{n}{} - \mean{}) - (\wght{}\zlatentn{n})\bigr)^{\trans}\covar{-1}\bigl((\xdatan{n}{} - \mean{}) - (\wght{}\zlatentn{n})\bigr) \biggr]
	\end{split}
	\label{eq: expanded loglkhd2}
\end{equation}

The log matrix identity $\ln\abs{aX} = D\ln(a) + \ln\abs{X}$can be used to expand and simplify the terms $\ln{\abs{(u_n\ident_d)}}$ and $\ln{\abs{u_n\covar{-1}}}$, giving


\begin{equation}
	\begin{split}
		\lkhd(\params) = & \ \dfrac{1}{N}\sum^N \biggl[ -D\ln{(2\pi)} + D\ln(u_n) - \hlf u_n \zlatentn{n}^{\trans}\zlatentn{n} + \dfrac{\nu}{2}\ln\left(\dfrac{\nu}{2}\right) \\ & \hdots - \ln\left(\Gamma\left(\dfrac{\nu}{2}\right)\right) + \dfrac{\nu}{2}\ln(u_n) - \dfrac{\nu}{2}u_n - \hlf\ln{\abs{\covar{}}}  \\ & \hdots - \hlf u_n (\xdatan{n}{} - \mean{})^{\trans}\covar{-1}(\xdatan{n}{} - \mean{}) + \hlf u_n (\xdatan{n}{} - \mean{})^{\trans}\covar{-1}(\wght{}\zlatentn{n}) \\ & \hdots + \hlf u_n (\wght{}\zlatentn{n})^{\trans}\covar{-1}(\xdatan{n}{} - \mean{}) - \hlf u_n (\wght{}\zlatentn{n})^{\trans}\covar{-1}(\wght{}\zlatentn{n})\biggr]
	\end{split}
	\label{eq: expanded loglkhd4}
\end{equation}

\subsection*{Q FUNCTION DERIVATION}
Given the definition of the log likelihood in Equation \ref{eq: expanded loglkhd4}, the Q-function and subsequently the EM update equations, can be derived. 

\begin{equation}
	\mathcal{Q} = \expect[{\zlatentn{},u|\xdatan{}{},\params}]{\lkhd(\params)} = \iint \lkhd(\params)p(\zlatentn{},u)\ \deriv{\zlatentn{}}\deriv{u}
\end{equation}

\begin{equation}
	\begin{split}
		\mathcal{Q} = & \ \dfrac{1}{N}\sum^N \biggl[ -D\ln{(2\pi)} + D\expect{\ln(u_n)} - \hlf \expect{u_n} \expect{\zlatentn{n}^{\trans}\zlatentn{n}} + \dfrac{\nu}{2}\ln\left(\dfrac{\nu}{2}\right) \\ & \hdots - \ln\left(\Gamma\left(\dfrac{\nu}{2}\right)\right) + \dfrac{\nu}{2}\expect{\ln(u_n)} - \dfrac{\nu}{2}\expect{u_n} - \hlf\ln{\abs{\covar{}}}  \\ & \hdots - \hlf \expect{u_n} (\xdatan{n}{} - \mean{})^{\trans}\covar{-1}(\xdatan{n}{} - \mean{}) + \hlf \expect{u_n} (\xdatan{n}{} - \mean{})^{\trans}\covar{-1}(\wght{}\expect{\zlatentn{n}}) \\ & \hdots + \hlf \expect{u_n} (\wght{}\expect{\zlatentn{n}})^{\trans}\covar{-1}(\xdatan{n}{} - \mean{}) - \hlf \expect{u_n(\wght{}\zlatentn{n})^{\trans}\covar{-1}(\wght{}\zlatentn{n})}\biggr]
	\end{split}
\end{equation}

Let the expectations of the latent parameters be denoted by
\begin{subequations}
	\begin{align}
		\expect{u_n} &= \bar{u}_n \\
		\expect{\ln(u_n)} &= \ln\tilde{u}_n \\
		\expect{\zlatentn{n}} &= \bar{\zlatentn{}}_n \\
		\expect{u_n\zlatentn{n}\zlatentn{n}^{\trans}} &= \bar{\mat{S}}_n
	\end{align}
\end{subequations}

Using the conditional distributions of the variables, exact solutions for the expectations can be recovered. 

\begin{align}
	p(\xdatan{n}{}|u_n) &= \int p(\xdatan{n}{}|\zlatentn{n},u_n)p(\zlatentn{n}|u_n)\deriv{\zlatentn{}}\\
	&= \int \gaussdistr{\wght{}\zlatentn{n}+\mean{}}{u_n^{-1}\covar{}}\gaussdistr{\vec{0}}{u_n^{-1}\ident_d}\deriv{\zlatentn{}}\\
	&= \gaussdistr{\mean{}}{u_n^{-1}\left(\covar{} + \wght{}\wght{\trans}\right)}
	\label{eq: x | u}
\end{align}

where $\mat{A} = \left(\covar{} + \wght{}\wght{\trans}\right)$, and 

\begin{equation}
	p(u_n|\xdatan{n}{}) \propto p(\xdatan{n}{}|u_n)p(u_n)
\end{equation}

\begin{equation}
	p(u_n|\xdatan{n}{}) = \gamdistr{u_n \bigg|\dfrac{D + \nu}{2}}{\dfrac{u_n(\xdatan{n}{} - \mean{})^{\trans}\mat{A}^{-1}(\xdatan{n}{} - \mean{})}{2} + \dfrac{\nu}{2}}
	\label{eq: u | x}
\end{equation}

Using the conditional distributions, values for the expectation of these variables, with respect to the other variables. 

\begin{equation}
	\bar{u}_n = \dfrac{\alpha}{\beta} = \dfrac{D + \nu}{ u_n(\xdatan{n}{} - \mean{})^{\trans}\mat{A}^{-1}(\xdatan{n}{} - \mean{}) + \nu}
	\label{eq: Gamma mean}
\end{equation}

is found using the equation for the mean of a Gamma distribution, and 

\begin{equation}
	\ln\tilde{u}_n = \psi\left(\dfrac{D + \nu}{2}\right) - \ln\left(\dfrac{u_n(\xdatan{n}{} - \mean{})^{\trans}\mat{A}^{-1}(\xdatan{n}{} - \mean{}) + \nu}{2}\right)
\end{equation}

is the expectation of the log Gamma distribution. 

\begin{equation}
	\zbarn = \mat{B}^{-1} \wght{\trans}\covar{-1}(\xdatan{n}{} - \mean{})
\end{equation}

where $\mat{B} = \wght{\trans}\covar{-1}\wght{} + \ident_d$ 

\begin{align}
	\bar{\mat{S}}_n &= \mat{B}^{-1} + \expect{u_n}\expect{\zlatentn{n}}\mathbb{E}[\zlatentn{n}^{\trans}]\\
	\bar{\mat{S}}_n &= \mat{B}^{-1} + \bar{u}_n\zbarn\zbarn^{\trans}
\end{align}

Expectation-maximisation update equations can be analytically found from the Q-function using 

\begin{equation}
	\params^{\mathrm{new}} = \argmax{\params}(Q(\params^{\mathrm{old}},\xdata{}))
\end{equation}

The EM update equations can be recovered by taking the derivative of the Q-function, with respect to the various parameters $\params$, and setting the resulting equations to zero and solving for the parameter in question. 





\subsubsection*{Covariance Update}

\begin{equation}
	\begin{split}
		\dfrac{\partderiv{\mathcal{Q}}}{\partderiv{\covar{-1}}} = & \dfrac{1}{N}\sum^N \biggl[ \hlf \covar{\trans} - \hlf \ubarn (\xdatan{n}{} - \mean{}) (\xdatan{n}{} - \mean{})^{\trans} + \hlf \ubarn (\xdatan{n}{} - \mean{})(\wght{}\zbarn)^{\trans} \\ & \hdots + \hlf \ubarn (\wght{}\zbarn)(\xdatan{n}{} - \mean{})^{\trans} + \hlf \ubarn \wght{} \bar{\mat{S}}_n \wght{\trans} \biggr]
	\end{split}
\end{equation}

setting $\dfrac{\partderiv{\mathcal{Q}}}{\partderiv{\covar{-1}}} = 0$ and rearranging, note that due to the symmetry of $\covar{}$, $\covar{\trans} = \covar{}$, the following update equation for the covariance can be defined. 

\begin{equation}
	\begin{split}
		\covar{} = & \dfrac{1}{N}\sum^N \biggl[\ubarn (\xdatan{n}{} - \mean{}) (\xdatan{n}{} - \mean{})^{\trans} - \ubarn (\xdatan{n}{} - \mean{})(\wght{}\zbarn)^{\trans} \\ & \hdots - \ubarn (\wght{}\zbarn)(\xdatan{n}{} - \mean{})^{\trans} + \ubarn \wght{} \bar{\mat{S}}_n \wght{\trans} \biggr]
	\end{split}
\end{equation}



\subsubsection*{Weight Update}

\begin{equation}
	\begin{split}
		\dfrac{\partderiv{\mathcal{Q}}}{\partderiv{\wght{}}} = & \dfrac{1}{N}\sum^N \biggl[\ubarn \covar{-1} (\xdatan{n}{} - \mean{})\zbarn^{\trans} - \covar{-1}\wght{}\bar{\mat{S}}_n \biggr]
	\end{split}
\end{equation}

letting $\dfrac{\partderiv{\mathcal{Q}}}{\partderiv{\wght{}}} = 0$ and rearranging,




\begin{equation}
	\wght{} = (\sum^N \ubarn(\xdatan{n}{} - \mean{})\zbarn^{\trans})(\sum^N \bar{\mat{S}}_n)^{-1} 
\end{equation}


\subsubsection*{Mean Update}

\begin{equation}
	\begin{split}
		\dfrac{\partderiv{\mathcal{Q}}}{\partderiv{\mean{}}} = \dfrac{1}{N}\sum^N \biggl[ \ubarn \covar{-1} (\xdatan{n}{} - \mean{}) - \ubarn \covar{-1} \wght{} \zbarn \biggr]
	\end{split}
\end{equation}

letting $\dfrac{\partderiv{\mathcal{Q}}}{\partderiv{\mean{}}} = 0$ and expanding,

\begin{equation}
	\begin{split}
		0 = \dfrac{1}{N}\sum^N \biggl[ \ubarn \covar{-1} \xdatan{n}{} - \ubarn \covar{-1}\mean{} - \ubarn \covar{-1} \wght{} \zbarn \biggr]
	\end{split}
\end{equation}

\begin{equation}
	\begin{split}
		\sum^N \biggl[\ubarn \covar{-1}\mean{}\biggr] = \sum^N \biggl[ \ubarn \covar{-1} \xdatan{n}{}  - \ubarn \covar{-1} \wght{} \zbarn \biggr]
	\end{split}
\end{equation}

premultiplying by $\covar{}$ and rearranging for $\mean{}$ gives

\begin{equation}
	\begin{split}
		\sum^N \biggl[\ubarn \mean{}\biggr] = \sum^N \biggl[ \ubarn (\xdatan{n}{}  - \wght{} \zbarn) \biggr]
	\end{split}
\end{equation}

\begin{equation}
	\begin{split}
		\ \mean{} = \dfrac{\overset{N}{\sum} \ubarn (\xdatan{n}{}  - \wght{} \zbarn)}{\overset{N}{\sum} \ubarn}
	\end{split}
\end{equation}

\subsubsection*{$\nu$ Update}

\begin{equation}
	\begin{split}
		\dfrac{\partderiv{\mathcal{Q}}}{\partderiv{\nu}} = \dfrac{1}{N}\sum^N \biggl[ \hlf\ln\left(\dfrac{\nu}{2}\right) + \hlf \cdot \dfrac{\nu}{2} \cdot \dfrac{2}{\nu} - \psi\left(\dfrac{\nu}{2}\right) + \hlf\ln(\tilde{u}_n) - \hlf\ubarn \biggr]
	\end{split}
\end{equation}

\begin{equation}
	\begin{split}
		\dfrac{\partderiv{\mathcal{Q}}}{\partderiv{\nu}} = \dfrac{1}{N}\sum^N \biggl[ \hlf  + \hlf\ln\left(\dfrac{\nu}{2}\right)  - \psi\left(\dfrac{\nu}{2}\right) + \hlf\ln(\tilde{u}_n) - \hlf\ubarn \biggr]
	\end{split}
\end{equation}

letting $\dfrac{\partderiv{\mathcal{Q}}}{\partderiv{\nu}} = 0$, multiplying by 2 and rearranging the sum,

\begin{equation}
	\begin{split}
		0 =  1  + \ln\left(\dfrac{\nu}{2}\right) - 2\ \psi\left(\dfrac{\nu}{2}\right) + \dfrac{1}{N}\sum^N \bigl[\ln(\tilde{u}_n) - \ubarn \bigr]
	\end{split}
\end{equation}

The maximum likelihood of $\nu$ can be found through solving th eabove equation using line search. 

\newpage
\section{Recovery of Rotation Matrix $\vec{R}$} \label{A: Rotation} 

The following theory for the recovery of the rotation matrix in Robust PCCA summarises the work presented by Archambeau et.al. in Appendix A of \cite{Archambeau2006} and the accompanying errata of Appendix A. for the same paper. 

Having defined
\begin{align}
	\vec{B}_1 &= {\wghtred{(1)}}^{\trans}{\covar{(1)}}^{-1}\wghtred{(1)} + \ident_d\\
	\vec{B}_2 &= {\wghtred{(2)}}^{\trans}{\covar{(2)}}^{-1}\wghtred{(2)}  + \ident_d 
\end{align}
The matrix $\mat{R}_1$ contains the eigenvectors of
\begin{equation}
	\vec{J}_1 = \bigl(\ident_d - \inv{\mat{B}_1}\bigr)^{\frac{1}{2}}\bigl(\ident_d - \inv{\mat{B}_2}\bigr)\bigl(\ident_d - \inv{\mat{B}_1}\bigr)^{\frac{1}{2}}
\end{equation}
with $\tilde{\Upsilon}^2$ corresponding eigenvalues. Similarly $\mat{R}_2$ contains the eigenvectors of 
\begin{equation}
	\vec{J}_2 = \bigl(\ident_d - \inv{\mat{B}_2}\bigr)^{\frac{1}{2}}\bigl(\ident_d - \inv{\mat{B}_1}\bigr)\bigl(\ident_d - \inv{\mat{B}_2}\bigr)^{\frac{1}{2}}
\end{equation}
with the same eigenvalues $\tilde{\Upsilon}^2$.

Given the above, it can finally be shown that the canonical directions are
\begin{align}
	\mat{U}^{(1)}_d &= {\mat{\Sigma}}^{-1}_{11}\wghtred{(1)}\bigl(\ident_d - \mat{B}_1^{-1} \bigr)^{-\frac{1}{2}} \mat{R}_1\\
	\mat{U}^{(2)}_d &= {\mat{\Sigma}}^{-1}_{22}	\wghtred{(2)}\bigl(\ident_d - \mat{B}_2^{-1} \bigr)^{-\frac{1}{2}} \mat{R}_2
\end{align}



\newpage
\section{Recovery of the modal properties from the state and output matrices.}
\label{A: Recovery of ModalProps}

Given a state matrix $\mat{A}$, the following eigenvalue problem 
\begin{equation}
	\mat{A}\vec{\psi} = \lambda\vec{\psi}
\end{equation}
can be solved to recover a set of eigenvalues $\lambda$ and corresponding eigenvectors $\vec{\psi}$. The respective eigenvalues correspond to a set of discrete time system poles. The continuous system poles can be recovered easily using
\begin{equation}
	\mu = \log (\Lambda)
\end{equation}
where $\Lambda$ is a diagonal matrix containing eigenvalues $\lambda$. Once recovered, the poles can be used to trivially calculate the natural frequencies in rad $s^{-1}$
\begin{equation}
	\omega_n = \frac{\abs{\mu}}{\Delta t} \ \ (\mathrm{rad} s^{-1})
\end{equation}
where $\Delta t$ is the sample time, and the damping ratios can be found using
\begin{equation}
	\zeta_n = -\mathrm{Re}\{\mu\}\oslash\abs{\mu}
\end{equation}
where $\oslash$ denotes a Hadamard (elementwise) division. 

Finally, the mode shape matrix $\Phi$ can also be recovered using the found output matrix $\mat{C}$ and the eigenvectors $\vec{\Psi}$, such that
\begin{equation}
	\Phi = \mathrm{Re}\{\mat{C} \vec{\Psi}\}
\end{equation}

\newpage
\setcounter{figure}{0}
\section{Varying Outlier Type Case Studies}\label{A: Varying Outlier Study}

Employing the same underlying linear dynamic system used in the outlier study shown in Section \ref{sec:Corrupted}, a range of alternative outlier cases were explored to observe the effects of varying outlier type on identification performance across both methods. 

\subsection{Periodic Block Dropout to Electrical Noise Floor}

\begin{figure}[H]
	\centering
	\includegraphics[width=0.7\textwidth]{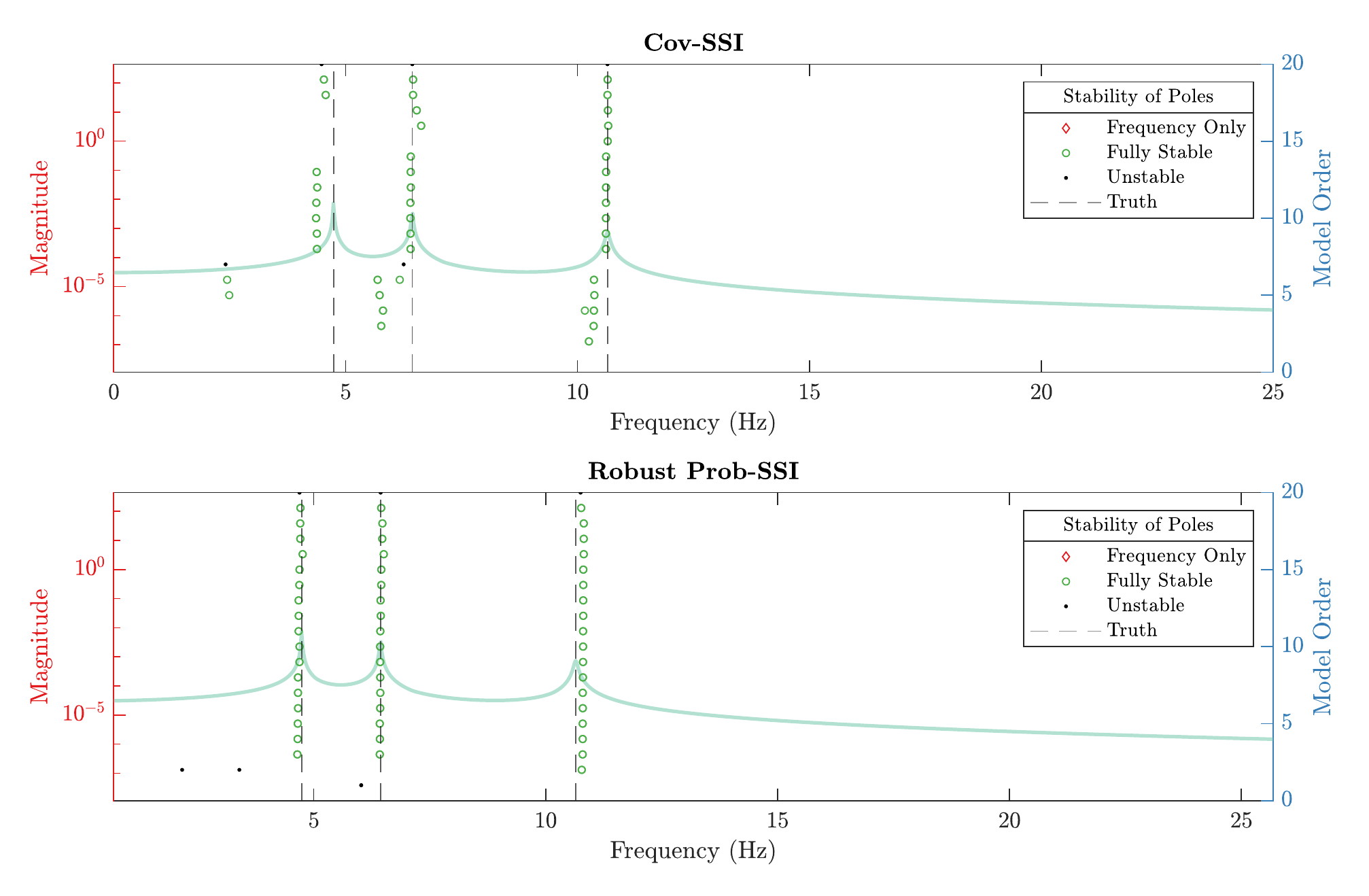}
	\caption{Consistency diagrams recovered using Cov-SSI (top) and robust Prob-SSI (bottom) for the periodic block dropout case.}
	\label{fig: consistency periodic}
\end{figure}

The outliers in this example are designed to mimic periodic dropout of a single sensor channel to a supposed `electrical noise floor', repeating in regular blocks with a duration of 0.01 seconds. Output response data and outliers are shown in Figure \ref{fig: response periodic}, whilst the consistency diagrams for Cov-SSI and robust Prob-SSI are shown in Figure \ref{fig: consistency periodic}.

\begin{figure}[H]
	\centering
	\includegraphics[width=0.7\textwidth]{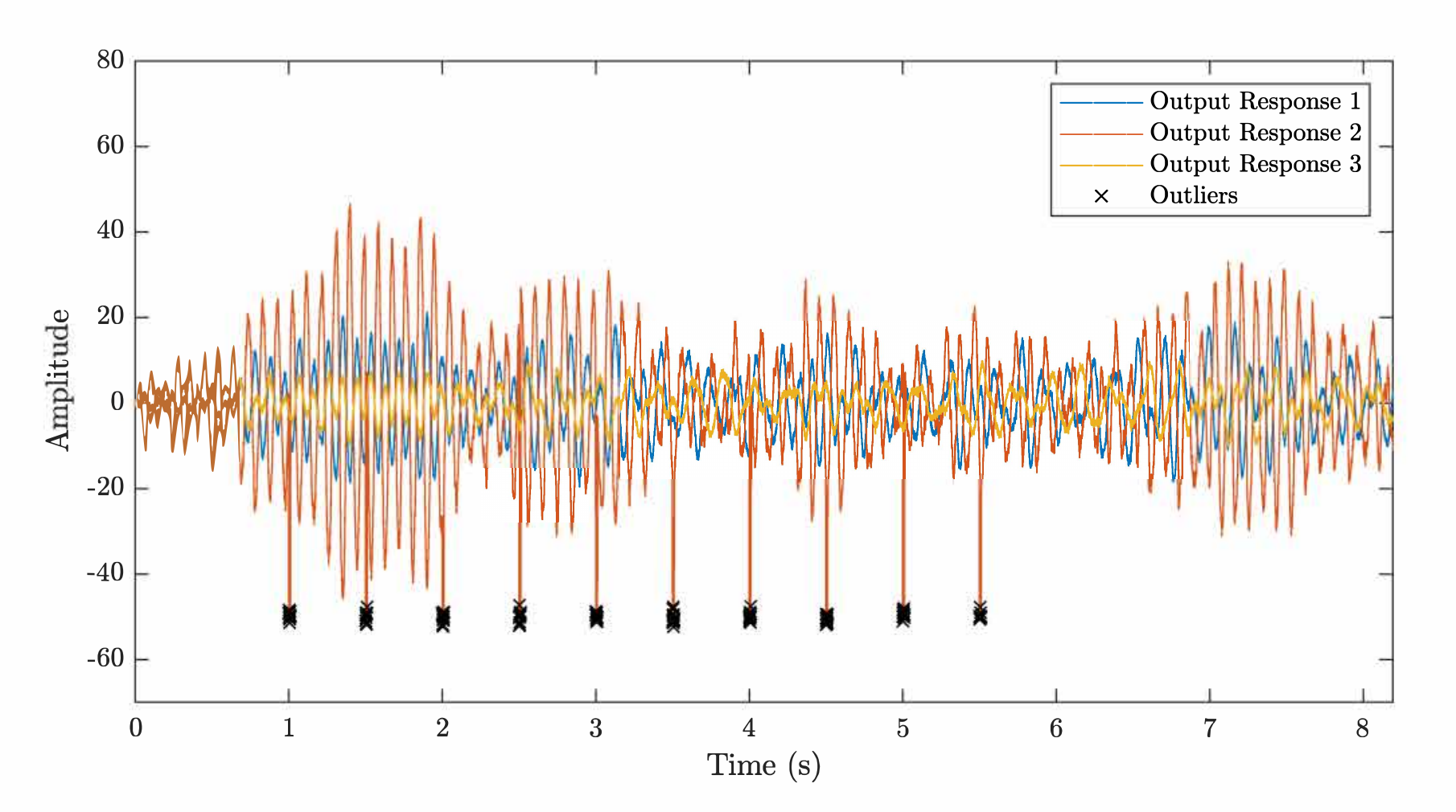}
	\caption{Response data from a simulated 3DOF `corrupted' dataset, containing 0.1\% of artificially introduced, periodic blocks of outliers in one channel. The value of the outliers was set to a specified value, plus some small amount of noise.}
	\label{fig:	response periodic}
\end{figure}

\newpage
\subsection{Clipping}

\begin{figure}[H]
	\centering
	\includegraphics[width=0.7\textwidth]{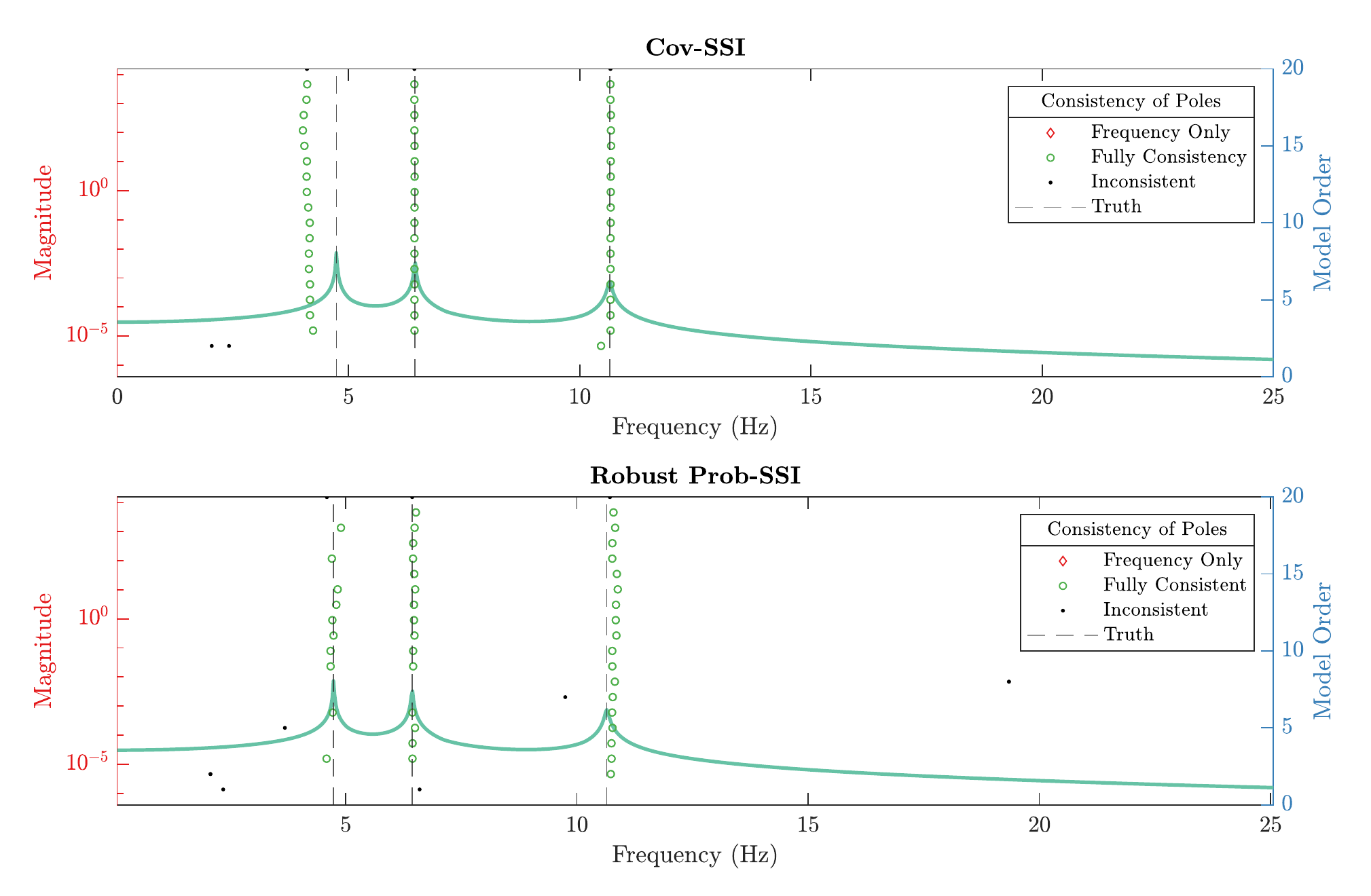}
	\caption{Consistency diagrams recovered using Cov-SSI (top) and robust Prob-SSI (bottom) using response data from a simulated 3DOF `corrupted' dataset, where all channels are clipped to 80\% of the maximum value of each channel.}
	\label{fig: consistency clipping}
\end{figure}

The outliers in this example are designed to mimic the clipping of all sensor channels to 80\% of the maximum amplitude of the individual signals. Output response data and the outliers are shown in Figure \ref{fig: response clipping}, whilst the consistency diagrams for Cov-SSI and robust Prob-SSI are shown in Figure \ref{fig: consistency clipping}.

\begin{figure}[H]
	\centering
	\includegraphics[width=0.7\textwidth]{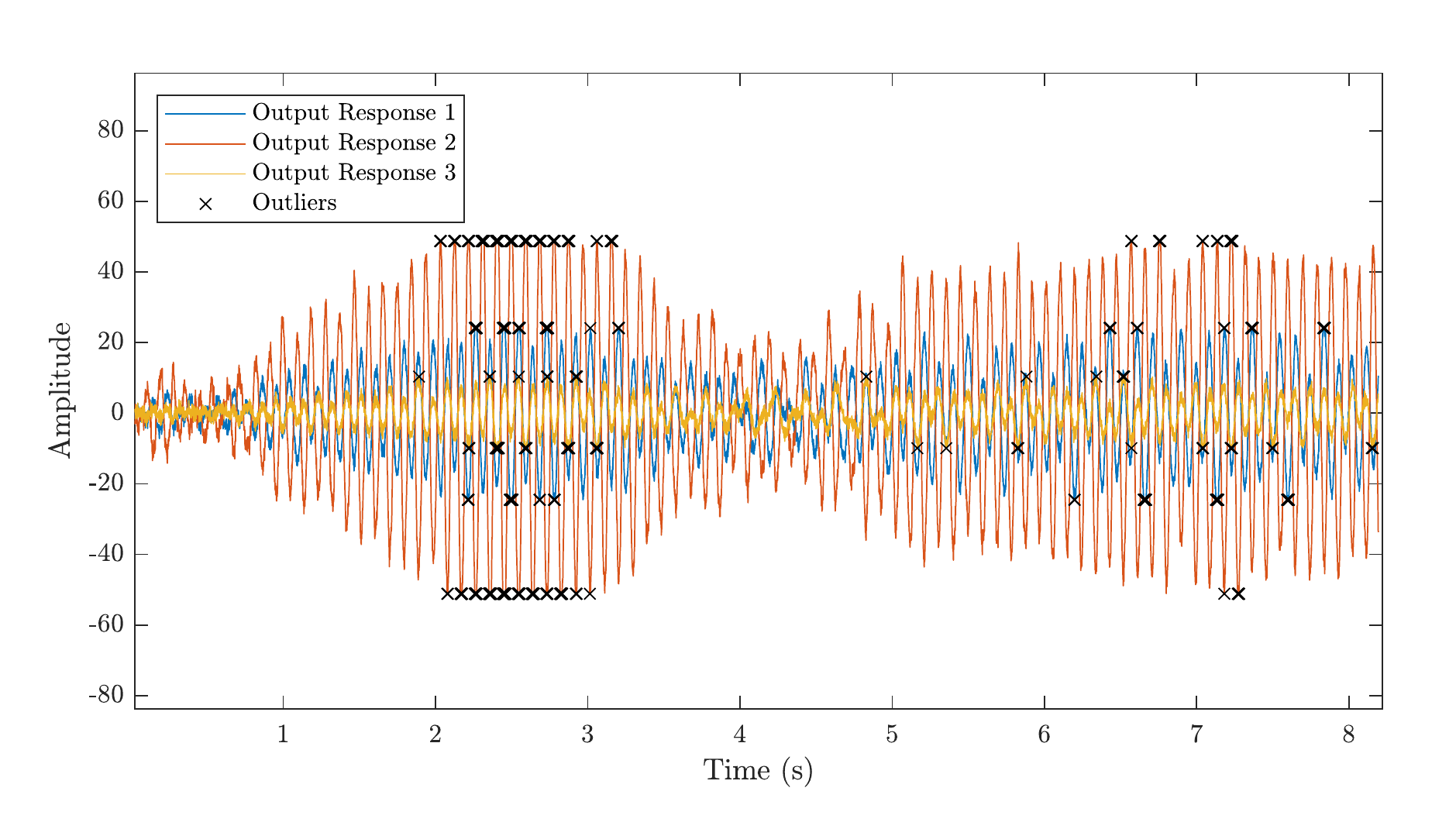}
	\caption{Response data from a simulated 3DOF `corrupted' dataset, where all channels were clipped to 80\% of the maximum value of each channel.}
	\label{fig:	response clipping}
\end{figure}

\newpage
\subsection{Block Dropout to Zero}

\begin{figure}[H]
	\centering
	\includegraphics[width=0.7\textwidth]{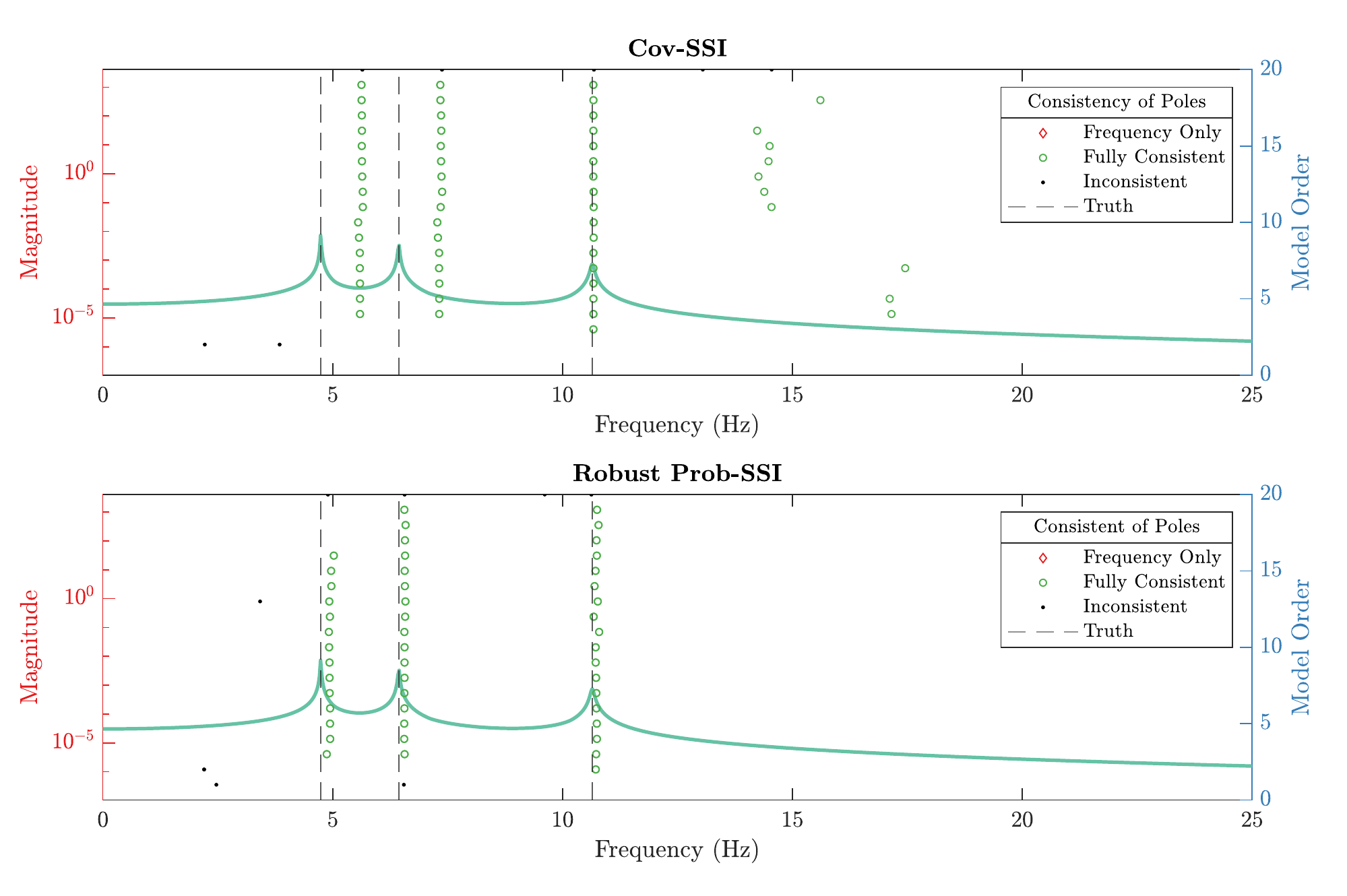}
	\caption{Consistency diagrams recovered using Cov-SSI (top) and robust Prob-SSI (bottom) using response data from a simulated 3DOF `corrupted' dataset, where values in a single channel were set to zero amplitude for a 1000 point block in the 8192 point long data series.}
	\label{fig: consistency zeroblock}
\end{figure}

The outliers in this example are designed to mimic a block dropout of a single sensor channel to zero. The dropout starts at 3 seconds and has a duration of 1 second. Output response data and outliers are shown in Figure \ref{fig: response zeroblock}, whilst the consistency diagrams for Cov-SSI and robust Prob-SSI are shown in Figure \ref{fig: consistency zeroblock}.

\begin{figure}[H]
	\centering
	\includegraphics[width=0.7\textwidth]{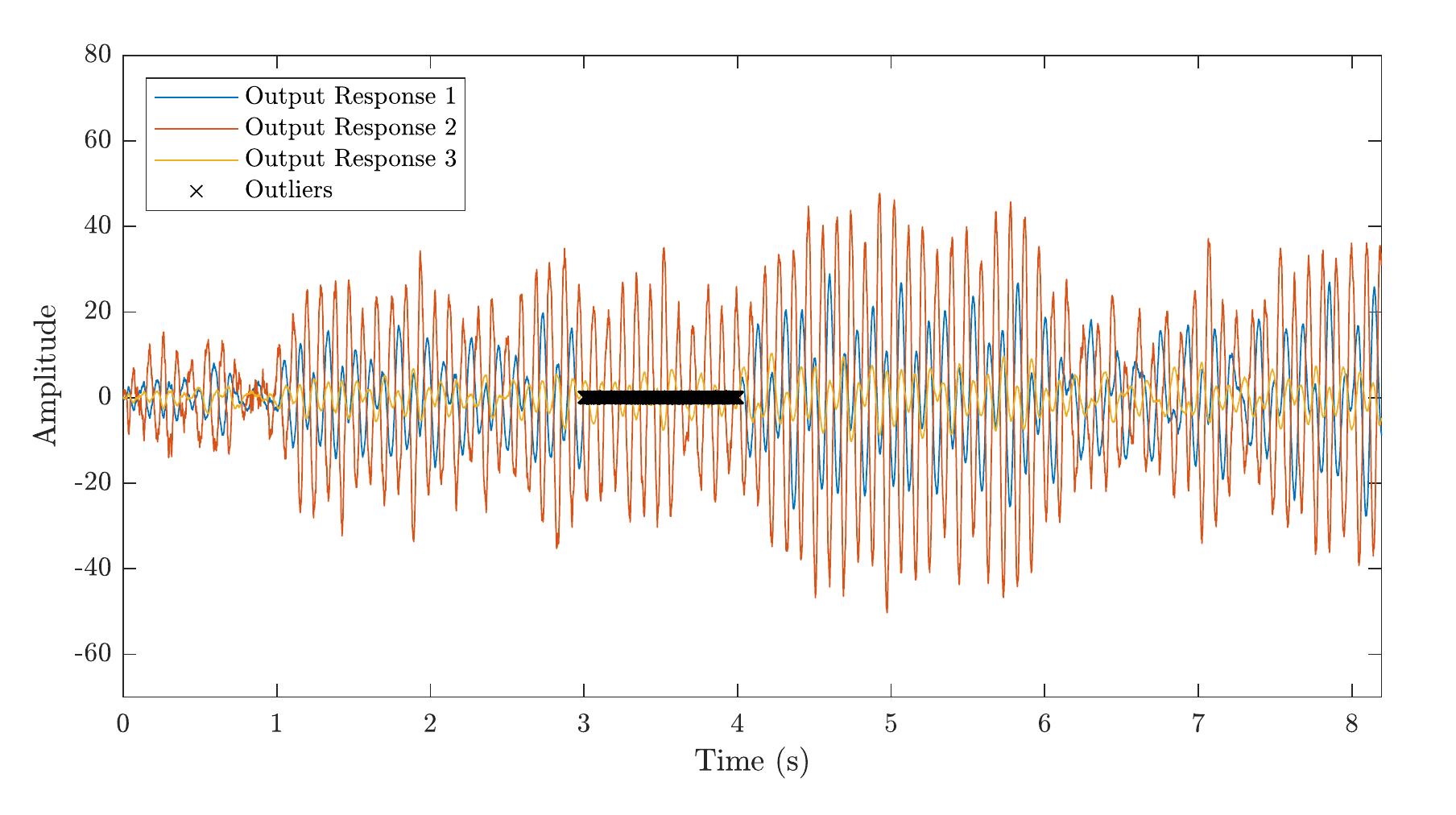}
	\caption{Response data from a simulated 3DOF `corrupted' dataset, where values in a single channel were set to zero amplitude for a 1000 point block in the 8192 point long data series.}
	\label{fig:	response zeroblock}
\end{figure}

%


%


\newpage
\section{Percentage Outlier Study}\label{A: Pct Outlier Study}

Employing the same parameters used in the outlier study shown in Section \ref{sec:Corrupted}, the effect of changing the percentage of outliers was explored to observe the effects on identification performance across both methods. 

\subsection{Random Outliers - 0.5\% }

\begin{figure}[H]
	\centering
	\includegraphics[width=0.7\textwidth]{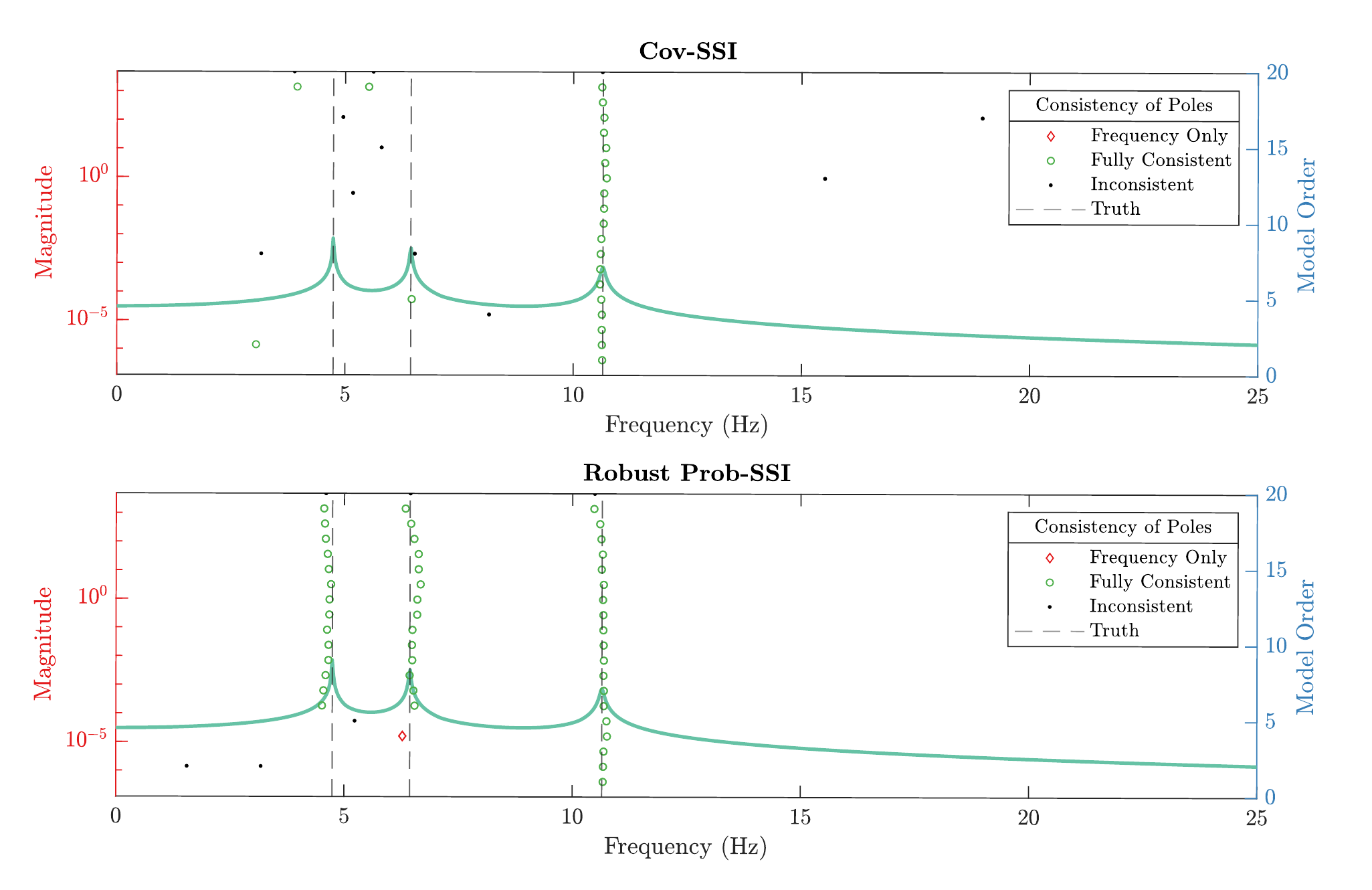}
	\caption{Consistency diagrams recovered using Cov-SSI (top) and robust Prob-SSI (bottom) using response data from a simulated 3DOF `corrupted' dataset, containing 0.5\% artificially introduced, randomly located outliers (in each channel), set to a specified value, plus some small amount of noise.}
	\label{fig:Stabil Diagrams Outliers 0.5}
\end{figure}



\newpage
\bibliographystyle{unsrtnat}
\bibliography{references.bib}

\end{document}

%% file: drg_continuity.tex
\usepackage{xparse}
\usepackage{mathtools}
\usepackage{amsthm,amsmath,amssymb}

\newcommand{\deriv}[1]{\text{d}#1}
\newcommand{\partderiv}[1]{\partial#1}
\renewcommand{\epsilon}{\varepsilon} 
\newcommand{\elmtreal}[1]{\in \mathbb{R}^{#1}}

\renewcommand{\vec}[1]{\boldsymbol{\mathbf{#1}}} 
\newcommand{\mat}[1]{\vec{\MakeUppercase{#1}}} 
\newcommand{\ident}{\mathbb{I}} 
\newcommand{\trans}{{\mathsf{T}}} 
\newcommand{\inv}[1]{#1^{-1}} 
\newcommand{\hlf}{\frac{1}{2}} 

\DeclarePairedDelimiter{\abs}{\lvert}{\rvert} 

\newcommand{\expect}[2][]{\mathbb{E}_{#1}\left[#2\right]}

\newcommand{\covar}[1]{\mat{\Sigma}^{#1}}

\newcommand{\argmax}[1]{\underset{#1}{\text{arg\,max}}}
\newcommand{\lkhd}{\mathcal{L}}
\newcommand{\mean}[1]{\vec{\mu}^{#1}}

\newcommand{\gaussdistr}[2]{\mathcal{N}\left(#1\,,#2\right)} 
\newcommand{\gamdistr}[2]{\mathcal{G}\left(#1\,,#2\right)} 

\newcommand{\xdatan}[2]{\vec{x}_{#1}^{{#2}}}
\newcommand{\zlatentn}[1]{\vec{z}_{#1}}
\newcommand{\zbarn}{\bar{\vec{z}}_{n}}
\newcommand{\ubarn}{\bar{u}_{n}}
\newcommand{\xdata}[1]{\mat{X}^{{#1}}}
\newcommand{\zlatent}{\mat{Z}}
\newcommand{\wght}[1]{\mat{W}^{#1}}
\newcommand{\wghtred}[1]{\mat{\breve{W}}^{#1}}
\newcommand{\wghtmle}[1]{\mat{\hat{W}}^{#1}}
\newcommand{\params}{\mat{\theta}}

\newcommand{\xstar}[1][]{
\ifthenelse{\equal{#1}{vec}}{\mat{X}_\star}{\vec{x}_\star}
}
\newcommand{\fstar}[1][]{
\ifthenelse{\equal{#1}{vec}}{\vec{f}_\star}{f_\star}
}
\newcommand{\ystar}[1][]{
\ifthenelse{\equal{#1}{vec}}{\vec{y}_\star}{y_\star}
}
\newcommand{\ypred}[1][]{
\ifthenelse{\equal{#1}{vec}}{\hat{\vec{y}}}{\hat{y}}
}
\newcommand{\ytrue}[1][]{
\ifthenelse{\equal{#1}{vec}}{\vec{y}}{y}
}

